# FINDINGS OF THE IWSLT 2024 EVALUATION CAMPAIGN

**Ibrahim Said Ahmad**[16]  **Antonios Anastasopoulos**[1]  **Ondřej Bojar**[4]  **Claudia Borg**[5]
**Marine Carpuat**[2]  **Roldano Cattoni**[3]  **Mauro Cettolo**[3]  **William Chen**[7]  **Qianqian Dong**[9]
**Marcello Federico**[8]  **Barry Haddow**[11]  **Dávid Javorský**[4]  **Mateusz Krubiński**[4]
**Tsz Kin Lam**[11]  **Xutai Ma**[6]  **Prashant Mathur**[8]  **Evgeny Matusov**[13]
**Chandresh Kumar Maurya**[20]  **John P. McCrae**[14]  **Kenton Murray**[10]  **Satoshi Nakamura**[12]
**Matteo Negri**[3]  **Jan Niehues**[15]  **Xing Niu**[8]  **Atul Kr. Ojha**[14]  **John E. Ortega**[16]
**Sara Papi**[3]  **Peter Polák**[4]  **Adam Pospíšil**[4]  **Pavel Pecina**[4]  **Elizabeth Salesky**[10]
**Nivedita Sethiya**[20]  **Balaram Sarkar**[20]  **Jiatong Shi**[7]  **Claytone Sikasote**[21]  **Matthias Sperber**[17]
**Sebastian Stüker**[18]  **Katsuhito Sudoh**[22,12]  **Brian Thompson**[8]  **Marco Turchi**[18]
**Alex Waibel**[7]  **Shinji Watanabe**[7]  **Patrick Wilken**[13]  **Petr Zemánek**[4]  **Rodolfo Zevallos**[19]

[1]GMU  [2]UMD  [3]FBK  [4]Charles U.  [5]U. Malta  [6]Meta  [7]CMU  [8]Amazon  [9]ByteDance
[10]JHU  [11]U. Edinburgh  [12]NAIST  [13]AppTek  [14]U. Galway  [15]KIT  [16]Northeastern U.
[17]Apple  [18]Zoom  [19]U. Pompeu Fabra  [20]IIT Indore  [21]U. Zambia  [22]Nara Women's U.

## Abstract

This paper reports on the shared tasks organized by the 21st IWSLT Conference. The shared tasks address 7 scientific challenges in spoken language translation: simultaneous and offline translation, automatic subtitling and dubbing, speech-to-speech translation, dialect and low-resource speech translation, and Indic languages. The shared tasks attracted 18 teams whose submissions are documented in 26 system papers. The growing interest towards spoken language translation is also witnessed by the constantly increasing number of shared task organizers and contributors to the overview paper, almost evenly distributed across industry and academia.

## 1 Introduction

The International Conference on Spoken Language Translation (IWSLT) is the premier annual scientific conference for all aspects of spoken language translation (SLT). IWSLT is organized by the Special Interest Group on Spoken Language Translation (SIGSLT), which is supported by ACL, ISCA and ELRA.

Like in all the previous 20 editions, this year's conference was preceded by an evaluation campaign featuring shared tasks addressing scientific challenges in SLT. This paper reports on the 2024 IWSLT Evaluation Campaign, which offered the following 7 shared tasks:

- **Offline SLT**, with focus on speech-to-text translation of recorded conferences and interviews from English to German, Japanese and Chinese.

- **Simultaneous SLT**, focusing on speech-to-text translation of streamed audio of conferences and interviews from English to German, Japanese and Chinese.

- **Automatic Subtitling**, with focus on speech-to-subtitle translation of audio-visual documents from English to German and Spanish and on compression of pregenerated German and Spanish subtitles.

- **Speech-to-speech Translation**, focusing on natural-speech to synthetic-speech translation of recorded utterances from English to Chinese.

- **Automatic Dubbing**, focusing on dubbing of production quality videos from English to Chinese.

- **Low-resource SLT**, focusing on the translation of recorded speech from Bhojpuri to Hindi, Irish to English, Marathi to Hindi, Maltese to English, North Levantine Arabic to English, Pashto to French, Tamasheq to French, Quechua to Spanish, and Bemba to English.

- **Indic Languages Track**, with focus on Speech-to-Text translation of TED talk audios from English to Indic languages including Hindi, Tamil, and Bengali.

The shared tasks attracted 18 teams (see Table 1) representing both academic and industrial organizations. The following sections report on each shared task in detail. Each section includes a description of the proposed challenge, the data and evaluation metrics used for training and testing systems, the received submissions, and finally a summary of the results. Detailed results for some of the shared tasks are reported in a corresponding appendix.

| Team | Organization |
|---|---|
| ALADAN | Vocapia, France, Lingea and Charles U., Czechia, Crowdee, Germany (Kheder et al., 2024) |
| APPTEK | Applications Technology (AppTek), Germany |
| CMU | Carnegie Mellon University, USA (Xu et al., 2024; Yan et al., 2024) |
| FBK | Fondazione Bruno Kessler, Italy (Papi et al., 2024; Gaido et al., 2024a) |
| HW-TSC | Huawei Translation Services Center, China (Wu et al., 2024; Wei et al., 2024) (Jiawei et al., 2024; Li et al., 2024a; Xie et al., 2024; Li et al., 2024b) |
| JHU | Johns Hopkins University, USA (Robinson et al., 2024) |
| KIT | Karlsruhe Institute of Technology, Germany (Koneru et al., 2024; Li et al., 2024c) |
| NAIST | Nara Institute of Science and Technology, Japan (Ko et al., 2024) |
| NICT | Nat. Inst. of Information and Comm. Technology, Japan (Dabre and Song, 2024) |
| NITKKR | National Institute of Technology Kurukshetra, India (Singh et al., 2024) |
| NYA | NetEase YiDun AI Lab, Hangzhou, China (Zhang et al., 2024) |
| QUESPA | Northeastern U, USA, U. de Pompeu Fabra, Spain, CMU, USA (Ortega et al., 2024) |
| RACAI | Romanian Academy, Romania (Gasan and Păiș, 2024) |
| SETU-DCU | SETech U, Ireland Unive di Pisa, Italy ADAPT, DCU, Ireland (Zafar et al., 2024) |
| UM, UOM | University of Malta, Malta (Nabhani et al., 2024; Abela et al., 2024) |
| UOM-DFKI | University of Malta, Malta, DFKI, Germany (Rishu et al., 2024) |
| BITSP | Birla Institute of Technology And Science - Pilani, India (Anand et al., 2024) |
| YMOSLEM | Independent Researcher, Ireland (Moslem, 2024) |

Table 1: List of participants to the IWSLT 2024 shared tasks

## 2 Offline SLT

Recent advances in deep learning are providing the opportunity to address traditional NLP tasks in new and completely different ways. One of these tasks is spoken language translation (SLT), an overarching problem that can be cast in various manners, ranging from offline to simultaneous processing, to produce either textual or speech outputs under both unconstrained and constrained conditions. This section reports on the 2024 round of the IWSLT Offline Speech Translation Track, which consists of translating audio speech from one language into text in a different target language without any specific time or structural constraints, different from the simultaneous (see §3), subtitling (§4), speech-to-speech (§5), and dubbing (§7) tasks. Under this general problem definition, the goal of the offline SLT track—the one with the longest tradition at IWSLT—is to continuously challenge this rapidly evolving technology by gradually introducing novel aspects that raise the difficulty bar.

### 2.1 Challenge

For years, SLT has been addressed by cascading an automatic speech recognition (ASR) system with a machine translation (MT) system. More recent trends involve using a single neural network to directly translate the input audio signal in one language into text in another language, bypassing intermediate symbolic representations such as transcriptions. In light of this evolution, the challenges addressed by the 2024 round of the offline track stem from the following considerations. **(1)** Although the results of the recent IWSLT campaigns have confirmed that the performance of end-to-end models is approaching that of cascade solutions, it is currently not clear which of the two technologies is more effective. Moreover, **(2)** all recent evaluations have been based on test sets extracted from TED talks, which represent a relatively simpler application scenario compared to the variety of potential deployments of SLT technology. In this controlled scenario, a single speaker delivers a prepared speech without background noise or interaction with other speakers. Finally, **(3)** last year's edition showed that introducing complexity to the scenario (e.g., including spontaneous speech, terminology, and dialogues) resulted in a clear performance degradation compared to using the classic TED talk test set.

Therefore, in addition to addressing the question of whether the cascade solution remains the dominant technology, this year we focused on understanding whether current state-of-the-art solutions can handle more complex scenarios (e.g., spontaneous speech, terminology, different accents, background noise, and dialogues). To shed light on these aspects, participants were challenged with data representative of different domains and conditions, namely:

- **TED Talks**[1] – the classic IWSLT evaluation material, for which fresh test data were collected also this year;

- **TV series** from ITV Studios[2] – data featuring multiple individuals interacting in various scenarios. The speech translation system needs to deal with overlapping speakers, different accents, and background noise;

- **Physical training videos** offered by Peloton[3] – data featuring individuals exercising in the gym. The speech translation system needs to deal with with background noise and an informal speaking style;

- **Accented English conversations** – data featuring conversations, each containing two friends interacting on a daily topic, such as hobbies and vacation. The speakers were selected to cover a wide range of English speakers around the globe. In addition to the variety of accents, another major challenge is the presence of spontaneous speech.

In continuity with the last two years, three language directions were proposed. Depending on the evaluation scenario, the language conditions covered are:

- English → German: TED talks, TV series, physical training videos, and accented English conversations;

- English → Japanese: TED talks.

- English → Chinese: TED talks.

---
[1] https://www.ted.com/
[2] https://www.itvstudios.com/
[3] https://www.onepeloton.com/

### 2.1.1 Test Suites

To further broaden the scope of evaluation conditions and explore specific aspects relevant to SLT, this year we provided participants with the option to submit additional test suites alongside the standard evaluation setting described above. The purpose of a test suite is to assess an SLT system on particular aspects that are generally hidden or overlooked by the classic evaluation frameworks. While the official evaluation relies solely on the designated official test sets, these supplementary test suites offer a valuable means to enhance system testing across a wider spectrum of phenomena. They also provide an opportunity to pinpoint specific and challenging issues that impact SLT performance. The particular test suite composition and its evaluation were fully delegated to the interested test suite provider.

## 2.2 Data and Metrics

**Training and development data.** Similar to the 2023 edition, participants were offered the possibility to submit systems built under three training data conditions:

1. **Constrained**: the allowed training data is limited to a medium-sized framework in order to keep the training time and resource requirements manageable. The complete list[4] of allowed training resources (speech, speech-to-text-parallel, text-parallel, text-monolingual) does not include any pre-trained language model.

2. **Constrained with large language models** (constrained$^{+LLM}$): in addition to all the constrained resources, a restricted selection[4] of large language models is allowed to give participants the possibility to leverage large language models and medium-sized resources. We reproduce the list of allowed LLMs in Table 2.

3. **Unconstrained**: any resource, pre-trained language models included, can be used with the exception of evaluation sets. This setup is proposed to allow the participation of teams equipped with high computational power and effective in-house solutions built on additional resources.

---
[4] See the IWSLT 2024 offline track web page: https://iwslt.org/2024/offline

| LLM | Source |
|---|---|
| Wav2vec 2.0 | https://github.com/pytorch/fairseq/blob/main/examples/wav2vec/README.md |
| Hubert | https://github.com/pytorch/fairseq/tree/main/examples/hubert |
| WavLM | https://github.com/microsoft/unilm/tree/master/wavlm |
| SpeechLM | https://github.com/microsoft/unilm/tree/master/speechlm |
| data2vec | https://github.com/facebookresearch/fairseq/tree/main/examples/data2vec |
| MBART | https://github.com/pytorch/fairseq/blob/main/examples/mbart/README.md |
| MBART50 | https://github.com/pytorch/fairseq/tree/main/examples/multilingual#mbart50-models |
| M2M100 | https://github.com/pytorch/fairseq/tree/main/examples/m2m_100 |
| Delta LM | https://github.com/microsoft/unilm/tree/master/deltalm |
| T5 | https://github.com/google-research/text-to-text-transfer-transformer |
| BLOOM (Note: only the small 560M parameter version) | https://huggingface.co/bigscience/bloom-560m#model-details |
| Mistral 7B Instruction Fine-tuned | https://huggingface.co/mistralai/Mistral-7B-Instruct-v0.1 |
| Mistral 7B Base Model | https://huggingface.co/mistralai/Mistral-7B-v0.1 |
| LLama2 7B Chat Model | https://huggingface.co/meta-llama/Llama-2-7b-chat-hf |
| Llama2 7B base model | https://huggingface.co/meta-llama/Llama-2-7b-hf |
| NLLB 3.3B | https://huggingface.co/facebook/nllb-200-distilled-1.3B |
| NLLB 1.3B | https://huggingface.co/facebook/nllb-200-3.3B |
| NLLB 600M | https://huggingface.co/facebook/nllb-200-distilled-600M |
| Seamless Models (SeamlessM4T/Streaming/Expressive) | https://github.com/facebookresearch/seamless_communication |

Table 2: List of LLMs allowed in the constrained$^{+LLM}$ training data condition.

The development data allowed under the constrained condition consists of the dev set from IWSLT 2010, as well as the test sets used for the 2010, 2013-2015 and 2018-2020 IWSLT campaigns. Besides this TED-derived material, additional development data were released to cover the three new scenarios included in this round of evaluation.

**Test data.** As in previous rounds of the offline track, the collection of new test data for the **TED talks** scenario started by isolating a set of talks (41 in total) that are not included in the current public release of MuST-C (Cattoni et al., 2021). Starting from this material, which was used to build the initial English-German test set, the talks for which Japanese and Chinese translations are available were selected to build the English-Japanese and English-Chinese test sets. Since further checks revealed a partial overlap between the selected talks and the TED2020 corpus[5] (Reimers and Gurevych, 2020) a final cleaning step had to be applied to remove the overlapping talks (4 for en-de, 4 for en-ja, none for en-zh). After this removal, the final test sets comprise 37 talks for English-German (corresponding to a total duration of 3h:07m:14s), 30 talks for English-Japanese (2h:14m:11s), and 30 talks for English-Chinese (3h:20m:19s).

For the **TV series** scenario, the 7 TV series for a total duration of 06h:01m are offered by ITV Studios.[6] Each series includes multiple speakers, background noise, and different audio conditions.

For the **Physical training** scenario, the 9 physical training videos for a total duration of 03h:59m are offered by Peloton.[7] Each video includes a single speaker in a room practicing sports activities with, often, background music and breathy voice.

For the **Accent challenge** scenario, the test set has 1,448 utterances that are sampled from 76 conversations in the Edinburgh International Accents of English Corpus (EdAcc, Sanabria et al., 2023). In total, the test set contains about 3.5 hours of audio data, 34k English words, 25.2k German words and 33 accents. The German translations are created from the English transcripts by our professional translators who are paid at a rate of 0.095 GBP per word. The translators, with access to the aligned audio files, were required to translate the transcripts in a fluent and faithful manner while allowing punctuation and casing. For example, hesitation tokens like "ACH" and "HMM" in the transcripts are not included in the translation. The complete translation guidelines are attached in Appendix B.1.

**Metrics.** Systems were evaluated with respect to their capability to produce translations similar to the target-language references. The similarity was measured in terms of multiple automatic met-rics: COMET[8] (Rei et al., 2020), BLEU[9] (Papineni et al., 2002a), chrF (Popović, 2015). Among them, this year COMET was chosen as the primary evaluation metric based the findings of Macháček et al. (2023) and Sperber et al. (2024), which indicate its highest correlation with human judgements. The submitted runs were therefore ranked based on the COMET calculated on the test set by using automatic resegmentation of the hypothesis based on the reference translation by mwerSegmenter,[10] using a detailed script accessible to participants.[11] Moreover, similar to last year's round, a human assessment was performed on the best-performing submission of each participant in order to enhance the soundness and completeness of the evaluation.

---
[5] https://opus.nlpl.eu/TED2020/en&de/v1/TED2020
[6] https://www.itvstudios.com
[7] https://www.onepeloton.com

### 2.3 Submissions

This year, 4 teams participated in the offline task, submitting a total of 38 runs. Table 3 provides a breakdown of the participation in each subtask showing, for each training data condition, the number of participants, the number of submitted runs and, for each training data condition (constrained, constrained$^{+LLM}$, unconstrained), the number of submitted runs obtained with cascade and direct systems. Notably, no direct system was submitted this year.

- CMU (Yan et al., 2024) participated with cascade en-de, en-ja, en-zh systems trained under the unconstrained condition. Their model consists of an ASR system based on Whisper and an MT system based on fine-tuned NLLB models. The ASR system is enhanced by the application of a specific fine-tuning to process unsegmented recordings without the need for a separate voice-activity detection stage. The MT systems generate a set of candidate translations via epsilon-sampling that are then pooled and the 1-best translation is selected using COMET-based Minimum Bayes-Risk decoding.

- HW-TSC (Wu et al., 2024) participated with cascade en-de, en-ja, en-zh systems trained under the constrained, constrained with Large Language Models, and unconstrained conditions. The authors used different training strategies for each different condition. Under the *constrained* condition, an ASR is trained from scratch testing Conformer and U2. All audio inputs are augmented with spectral augmentation), and Connectionist Temporal Classification (CTC) is added to make the model converge better. The MT system takes advantage of the Deep Transformer-Big model structure, R-Drop and data selection to identify in-domain data from a large pool of parallel data. Under the *constrained + LLM* condition, the ASR system is a combination of the wav2vec2 encoder and mBART50 decoder, where the self-attention of the encoder and decoder are frozen and all constrained are used for fine-tuning. The MT system is based on Llama2-7B fine-tuned with parallel data and source language consistent instructions, and applying CPO. Under the *unconstrained* condition, the ASR system is based Whisper fine-tuned and MuST-C, while the MT model selects the 1-best translation from a pool of candidates generated both with NMT and LLM using COMET. Audio segmentation is performed using SHAS.

- KIT (Koneru et al., 2024) participated with a cascade en-de system trained under the constrained with Large Language Models condition. This submission is based on a four-step approach. The audio is first transcribed by a fine-tuned ASR, the n-best list is then processed by an LLM to generate the best hypothesis. The final transcripts is translated to generate the text in the target language. The transcript and the translation are then paired and document- level automatic post-editing is applied to improve the coherence of the translations. The ASR is based on the combination of WavLM encoder and mBART50 decoder fine-tuned on the task data. Audio segmentation is based on SHAS, but a longformer technique is also tested to use context better. The ASR refiner and the MT post-editor are fine-tuned versions of Mistral 7B Instruction-tuned LLM using QLoRA, while NLLB 200 3.3B is used as the MT system. The post-editing step showed to be less effective when the ASR quality is low. For this reason, LLM refinement is not used for the EPTV and ITV datasets.

---
[8] Unbabel/wmt22-comet-da
[9] BLEU+case.mixed+numrefs.1+smooth.exp+tok.13a +version.1.4.14
[10] https://www-i6.informatik.rwth-aachen. de/web/Software/mwerSegmenter.tar.gz
[11] https://github.com/isl-mt/SLT.KIT/blob /master/scripts/evaluate/Eval.sh

| English-German | | | | | | | | |
|---|---|---|---|---|---|---|---|---|
| Participants | Runs | Constrained | | Constrained$^{+LLM}$ | | Unconstrained | | |
| 4 | 14 | 2 | Cascade | 2 | 3 | Cascade | 3 | 9 | Cascade | 9 |
| | | | Direct | - | | Direct | - | | Direct | - |
| English-Chinese | | | | | | | | |
| Participants | Runs | Constrained | | Constrained$^{+LLM}$ | | Unconstrained | | |
| 3 | 13 | 2 | Cascade | 2 | 2 | Cascade | 2 | 9 | Cascade | 9 |
| | | | Direct | - | | Direct | - | | Direct | - |
| English-Japanese | | | | | | | | |
| Participants | Runs | Constrained | | Constrained$^{+LLM}$ | | Unconstrained | | |
| 3 | 11 | 2 | Cascade | 2 | 2 | Cascade | 2 | 4 | Cascade | 4 |
| | | | Direct | - | | Direct | - | | Direct | - |

Table 3: Breakdown of the participation in each sub-task (English→German, English→Chinese, English→Japanese) of the IWSLT offline ST track. For each language direction, we report the number of participants, the number of submitted runs and, for each training data condition (constrained, constrained$^{+LLM}$, unconstrained), the number of submitted runs obtained with cascade and direct systems.

- NYA (Zhang et al., 2024) participated with cascade en-de, en-ja, en-zh systems trained under the unconstrained condition. The ASR is based on Whisper-v3-large, while the MT system is a wider and deeper Transformer model. The MT model is enhanced by leveraging several techniques such as R-Drop, data augmentation with backward translations, domain adaptation via data filtering, and ASR output adaptation where the human-quality transcript in the SLT data is replaced with the automatic transcript. The final MT model is an ensemble of two/three models. The audio is segmented using SHAS.

### 2.4 Results

We will analyse the different aspects of the results by language pair.

#### 2.4.1 English to German

**Correlation between BLEU, COMET and DA scores** Table 25 shows the aggregated result of the participated systems on the four test sets. In terms of ranking based on the BLEU score, NYA wins 3 out of 4 test sets, except on ITV which CMU and HW-TSC(U) have a tie. However, the ranking is substantially changed when COMET is used. In this case, CMU is the winning system in all conditions, indicating that this submission achieves the best performance. But in contrast to last year when the human evaluation validated the automatic metric rankings, the correlation between the automatic rankings and the human ranking is not as good as shown in Table 18. (More details on our human evaluation using DA are provided in Appendix A.2.1.) For the human evaluation, HW-TSC(C+) achieves the best performance overall and has the best DA ranking on 3 out of four test sets. Only on the accent test set, NYA has better scores. However, it is worth noticing that no system performs significantly better than HW-TSC(C+) on any dataset.

The results show that it is essential to perform a human evaluation since no automatic metric, at the moment, can predict the performance of the individual systems well. Furthermore, additional research on performing reliable automatic metrics for speech translation would be very valuable.

It is interesting to note that all the submissions are based on the cascade architecture this year. This is an important change compared to previous editions where the end-to-end architectures competed with the cascade ones.

**Context Beyond Segment Level** One of the participating teams, KIT, used document-level post-editing to improve the coherence of translation. We note that while document-level consistency is a critical feature of text and speech translation, our evaluation this year does not reflect it yet. All used automatic metrics are segment-oriented. As detailed in Appendix A.2.1 also the particular setup of DA this year did not allow the annotators to consider longer context because the segments were shuffled for DA. (Two neighbouring segments were provided but only to account for segmentation errors, not for assessment of context-level phenomena.) It is therefore conceivable that the outputs of the KIT system were somewhat penalized.

**Domains** Similar to last year's edition, we evaluated each submitted system on different domains. First of all, the results show that the systems perform very differently in the different domains. When looking at the human ranking, the best quality is achieved in the TED domain. This is not surprising, since research has focused on this for many years and a significant amount of training resources exist. The performance on ITV and Peloton is lower, and the Accent data set appears to be the most challenging condition, indicating that speech translation remains an unsolved problem.

The availability of human rankings of the same systems across different domains allows us also to analyse whether automatic scores can be used to assess the quality of SLT system across domains. When ranking the difficulty level of different domains, we see that COMET ranks them similar than the human ranking except that COMET shows overall lower scores for Peloton, identifying Peloton as more challenging than Accent. In constrast, string-based metrics like BLEU are not able to do this. This also shows that additional metrics might be needed to measure the quality across domains.

**Data conditions** On top of the above, we can also observe the improvement in both BLEU and COMET scores caused by using an additional large language model or additional data. HW-TSC submitted three primary systems for each data condition, and both the unconstrained (U) and the constrained$^{+LLM}$ (C$^+$) models have a noticeable gain over the constrained model (C). The two better models perform similarly in both BLEU and COMET. Interestingly, additional training data beyond the language model data does not significantly improve. In terms of DA score, the constrained$^{+LLM}$ model is >0.6 points better than the other models in different data conditions.

**Progress compared to last year** We also performed an automatic evaluation of the system on the test sets from last year from the domains TED, EMPAC, and ACL. The results are summarized in Table 26. Although the participants optimized for different domains, for each domain and each metric this year's submissions achieved the best performance. When comparing the best submission from this year and last year, this year's submission is between 4.4 and 1.5 BLEU points better and 1.1 to 2.7 COMET percent points better than the best system from last year.

**Performance by accents** For the accent test set, we performed an additional details analysis for the different accents.

Figure 1 shows the BLEU and COMET of each system across the 33 accents. The numbers in parenthesis are audio duration in the format of "minutes:seconds". We use the self-reported labels from the original work as the prior choice for accent labeling. Since accents could be loosely defined (e.g., multi-class), subjective, and most speakers in the annotation are not the related experts, we thus derive the labels from other attributes, such as the first language of the speaker, if necessary and refine the labels to country-level. There is one speaker who declares his accent as "Trans-Atlantic" and speaks multiple first languages. We assign this special case as "Mixed".

The aggregated result on Table 25 shows that CMU is the winning system on Accent when COMET is used for ranking, whereas NYA would be the winner if BLEU is used instead. Does this winning situation occurs on a wide range of accents or on a small subset? The breakdown on Figure 1 shows that CMU (the blue-diamond points) has better COMET scores, especially relative to NYA, and is within Top-2 on a wide range of accents. Similar observations are found in the better BLEU scores of NYA (the yellow-star points).

For the three primary systems submitted by HW-TSC (the red points), their performances are rather consistent across the 2 metrics and the accents. In most cases, both the constrained$^{+LLM}$ (the circles) and the unconstrained models (the squares) perform similarly, while the constrained model (the triangles) falls slightly behind. In the North Macedonian and the Pakistani accents, the constrained model seems to be better in both BLEU and COMET, but their data sizes are rather small, i.e. <1 minute. In the constrained LLM setting, the HW-TSC system in general performs better than the KIT system in a wide range of accents, but the KIT system has a slight edge in Indonesian,

Israeli and Japanese accents.

The macro-average across accents are 18.7 BLEU and 0.679 COMET. Despite their fairly large test sizes, French, Irish, Jamaican, Kenyan and Vietnamese are below average. In Brazilian, German, Mexican and South African accents, all systems perform rather poorly, i.e., <10 BLEU. Potential causes are the train-test mismatch in accents, their small test sizes and the re-segmentation error in the short utterances. Additionally, these speeches contain a mix of disfluencies and named entities, e.g., food ingredients, imposing further translation challenges.

#### 2.4.2 English to Japanese

For the English to Japanese direction, we only have one test condition, the TED domain. In this case, the HW-TSC is the winner in all metrics, BLEU, COMET, and human ranking. However, the order of the submissions from HW-TSC varies across different metrics. Furthermore, the other two participants perform similarly on human ranking, but CMU is clearly better on COMET and NYA is clearly better on BLEU. This again suggests that the automatic metrics do not perform sufficiently well on speech translation tasks yet. Similar to the En-De language direction, all the submitted systems are based on the cascade architecture.

When comparing the submissions from this year and last year on the two progress test sets (TED and ACL), we again see a clear improvement compared to last year's best systems.

For the data conditions, we see again a better performance of the unconstrained (U) and the constrained$^{+LLM}$ (C$^+$) submissions from HW-TSC compared to the system using only constrained data. However, this does not hold for the BLEU metric and the human evaluation. In these metrics, we see no clear benefit from using more data.

#### 2.4.3 English to Chinese

For the English to Chinese direction, we also have only one test condition, the TED domain. In this case, the HW-TSC is the best system in human evaluation and COMET, while NYA performed best in BLEU. While this could indicate a good correlation between human evaluation and COMET, NYA actually serves as a counterexample: it performed worst in COMET and second best in human evaluation. This again suggests that the automatic metrics do not work reliably on speech translation tasks yet. Similar to the other language directions, all the submitted systems are based on the cascade architecture.

When comparing the submissions from this year and last year on the two progress test sets (TED and ACL), we again see a clear improvement compared to the best systems of last year.

For the data conditions, we see again a better performance of the unconstrained (U) and the constrained$^{+LLM}$ (C$^+$) submissions from HW-TSC compared to the system using only constrained data, when considering the COMET metric and the human evaluation.

## 3 Simultaneous SLT

Simultaneous speech translation focuses on translating speech in real-time, in manner vaguely similar to simultaneous interpreting. The system is designed to begin translating before the speaker has finished their sentence. This technology is particularly useful in scenarios such as international conferences, personal travel, or public emergency events.

This year, the task included two tracks: speech-to-text and speech-to-speech, covering four language directions: English to German, English to Chinese, English to Japanese, and Czech to English—a new language direction added this year.

### 3.1 Challenge

We have retained the settings from last year's shared task. A single latency constraint is introduced for each of the tracks:

- An average lagging of 2 seconds for the speech-to-text track.

- A starting offset of 2.5 seconds for the speech-to-speech track.

Participants are allowed to submit no more than one system per track and language direction, provided the system's latency remains within the specified constraints. The latency performance of the systems is evaluated using the open MuST-C tst-COMMON test set (Di Gangi et al., 2019). Submissions were accepted only in the form of Docker images, which were later executed by the organizers on the blind-test set in a controlled environment. An example implementation was provided using the SimulEval toolkit (Ma et al., 2020).

| set | domain | #utter. | #words/ utter. | duration (min) |
|---|---|---|---|---|
| dev | ParCzech | 276 | 24 | 56 |
|  | ELITR | 314 | 13 | 28.6 |
| test | MockConf | 1113 | 14 | 129.5 |

Table 4: Statistics of the dev and test sets for the Czech-English simultaneous task.

### 3.2 Data

To simplify the setting and allow participants to focus on the new modeling aspects of simultaneous translation, we adhere to the constraints with large language models as defined for the offline SLT task, see Section 2.2 above. This is the sole data condition for the task. The test data differ across different language pairs:

**English to German, Chinese, and Japanese** Common TED Talks, which are the same as those used in the Offline task, as described in Section 2.2.

**Czech to English** The devset was created from two sources:

- A subset called "context" was taken from ParCzech 3.0 (Kopp et al., 2021), consisting of consecutive recordings of Parliament of the Czech Republic.

- An entire recording of a debate about AI from the ELITR test set (Ansari et al., 2021).[12]

The reference translations of the devset were done by students of translation studies from the Faculty of Arts at Charles University.

The testset was gathered from mock conferences that were part of the interpreting curriculum of the Faculty of Arts at Charles University. A speaker pretends to be a celebrity or an interesting person and delivers a made-up speech on a pre-determined topic. We included 13 such speeches. The reference translations were provided by professional translators. Due to confidentiality of recordings, the testset is not released to the community. The statistics of the data are displayed in Table 4.

---
[12] https://github.com/ELITR/elitr-testset /tree/master/documents/2021-theaitre-r elated/robothon-debate

### 3.3 Evaluation

We evaluate two aspects of the model: quality and latency.

**Quality** We conducted both automatic and human evaluation. BLEU score (Papineni et al., 2002b) is used for automatic quality evaluation. For speech output, the BLEU score is computed on the transcripts from Whisper (Radford et al., 2023) ASR model. The ranking of the submission is based on the BLEU score on the Common blind test set. The human evaluation was conducted in English-to-German/Chinese/Japanese, as described in A.1.

**Latency** We only conducted automatic evaluation. We report the following metrics for each speech-to-text systems.

- Average Lagging (AL; Ma et al., 2019)

- Length Adaptive Average Lagging (LAAL; Polák et al., 2022; Papi et al., 2022a)

- Average Token Delay (ATD; Kano et al., 2023)

- Differentiable Average Lagging (DAL; Arivazhagan et al., 2019)

For speech-to-speech systems, we report start-offset, end-offset and Average Token Delay. The latency metrics will not be used for ranking.

### 3.4 Submissions

Four teams in total submitted systems this year, with all teams participating in at least one language direction in the speech-to-text track. All teams entered the English-to-German track; three teams entered the English-to-Chinese and English-to-Japanese tracks; and two teams entered the Czech-to-English track, to which we added a Whisper-based benchmark. For the speech-to-speech track, two teams submitted systems, with one team submitting for all language directions and the other only in the English-to-Japanese direction.

**CMU (Xu et al., 2024)** participated in the speech-to-text track for the English-to-German direction. Their system integrates the WavLM-based speech encoder (Chen et al., 2021), a modality adapter, and the Llama2-7B-based decoder (Touvron et al., 2023). The training is conducted in two stages: modality alignment and full fine-tuning, both performed on MuST-C v2 data (Cattoni et al., 2021). The two-stage training results in an offline speech translation model, which is then adapted to a simultaneous speech translation model with a simple fixed hold-n policy.

**FBK (Papi et al., 2024)** participated in all language directions of the speech-to-text track. Their system is a unified multilingual simultaneous speech translation system, combining AlignAtt (Papi et al., 2023b) and SeamlessM4T-medium (Seamless Communication et al., 2023). The SeamlessM4T model is directly used in its streaming mode without additional retraining. The generated hypotheses are further processed through AlignAtt for policy learning. Based on diverse training sources, the model can translate into approximately 200 target languages from 143 source languages.

**HW-TSC (Li et al., 2024a)** participated in all language directions of both the speech-to-text and speech-to-speech tracks. Except for the Czech-to-English direction, all other models utilize cascaded simultaneous speech translation approaches by combining offline speech recognition, machine translation, and text-to-speech. For the Czech-to-English direction, they utilize the offline SeamlessM4T (Seamless Communication et al., 2023) as the backbone for speech-to-text translation, combined with a text-to-speech system. They followed their last year's submissions as the base setting (Guo et al., 2023). Additionally, they applied online voice-activity-detection-oriented segmentation, chunk padding in the speech recognition system to achieve smaller delays, and added an ensemble strategy for machine translation to achieve better stability. For end-to-end speech-to-text translation, they fine-tuned the SeamlessM4T model using the suggested data in the simultaneous SLT shared task.

**NAIST (Ko et al., 2024)** participated in three language directions of the speech-to-text track. Their speech-to-text system combined HuBERT (Hsu et al., 2021) and mBART (Liu et al., 2020b) in an end-to-end fashion, with a local agreement policy (Liu et al., 2020a; Polák et al., 2022). Their speech-to-speech system further applied an incremental text-to-speech module tuned with AlignAtt policy (Papi et al., 2023b).

**ORGANIZER'S BENCHMARK** by Charles University was prepared for the Czech-to-English direction. The system is based on Whisper (Radford et al., 2023) version `large-v2`. We applied an onlinization technique (Polák et al., 2022, 2023a,b) to utilize the offline Whisper model in the simultaneous regime, and applied prompting to leverage the translation history from previous segments. Due to organizational reasons, the benchmark was run on different hardware so the comparison of computationally-aware latency with other systems is not possible.

### 3.5 Results

We rank the system performance based on BLEU scores. The detailed results can be found in the respective tables in Appendix A.2.3.

**Speech-to-Text** The ranking of the speech-to-text track is as follow

- English to German (Table 29): HW-TSC, CMU, NAIST, FBK

- English to Chinese (Table 30): HW-TSC, NAIST, FBK

- English to Japanese (Table 31): HW-TSC, NAIST, FBK

- Czech to English (Table 32): ORGANIZER'S BENCHMARK (with context of 2 segments), FBK, HW-TSC

**Speech-to-Speech** As mentioned in Section 3.4, two teams submitted speech-to-speech track this year. HW-TSC submitted systems on all language directions and NAIST submitted on English to Japanese Direction. We only rank the English to Japanese Direction. The rank is: HW-TSC, NAIST. See Table 33 for more details.

### 3.6 Conclusions

Over the past four years, the IWSLT has consistently featured simultaneous translation tasks, reflecting a growing interest and impressive progress in this area. The shared task also brings the establishment of standardized evaluation protocols for simultaneous translation research. The recent integration of foundation models has further expanded the potential of this task. All teams integrated such models into their submissions using different approaches. CMU and NAIST teams combined two foundation models each specialized in one modality (speech encoder and text decoder) together using fine-tuning, while others chose existing ST models such as SeamlessM4T or Whisper and modified them for simultaneous use. Surprisingly, even large models (e.g., the CMU's Llama2-7B-based decoder) achieved competitive computationally-aware latencies.

The only cascaded system in the competition (HW-TSC) was consistently rated first in three language pairs. Nevertheless, according to all latency measurements, this system also exhibited the highest computationally-aware latencies.

One of the interesting points this year is the newly-added Czech-to-English translation direction where we included our Whisper-based benchmark. When operating at the segment level, this benchmark performed worse than participants' systems, but given one or two of its previous translation outputs, it improved over them. This confirms that the role of context is very important in speech translation task and the best uses of LLMs for this task are still to be found.

Several promising directions for future improvements remain. Investigating downstream tasks such as cross-lingual dialogues could provide deeper insights into practical applications of simultaneous translation. Developing more interactive evaluation methods could enhance the understanding and effectiveness of these systems. Lastly, optimizing the evaluation procedure to expedite the process remains crucial, as the current system managed by the organizers can be time-consuming.

## 4 Automatic Subtitling

In recent years, the task of automatically creating subtitles for audiovisual content in another language has gained a lot of attention due to the rapid increase in the global distribution and streaming of movies, series, and user-generated videos. Reflecting these trends, the automatic subtitling track was introduced for the first time in 2023 as part of the IWSLT Evaluation Campaigns. Given the growing interest in this area, the task has been continued this year with the addition of a new sub-track, **subtitle compression**, alongside the existing **automatic subtitling** sub-task from the previous edition.

In the automatic subtitling task, participants were asked to generate subtitles in German and/or Spanish from English speech in audiovisual documents. In the new subtitle compression task, participants were required to automatically rephrase subtitles that did not comply with the reading speed constraint (i.e., subtitles exceeding a certain length/time ratio given in characters per second) to ensure they met the required standards.

The decision to have works focusing on this specific aspect of subtitling is highly motivated by the existing requirements posed by subtitles providers (Papi et al., 2023a). In fact, the constraint on the reading speed is a commonly adopted standard to ensure that viewers can enjoy audiovisual content without experiencing fatigue or distraction due to excessive reading demands (Kruger, 2001). Therefore, adhering to this limit is crucial, making the development of ad-hoc methods to improve automatically generated subtitles that exceed this threshold of particular interest.

### 4.1 Challenge

**Automatic Subtitling.** The task of automatic subtitling is multifaceted: starting from speech, not only must the translation be generated, but it must also be segmented into subtitles that comply with constraints ensuring a high-quality user experience. These constraints include proper reading speed, synchrony with the voices, the maximum number of subtitle lines, and characters per line. Most audio-visual companies define their own subtitling guidelines, which can slightly differ from each other. In the case of IWSLT participants, we asked to generate subtitles according to specific guidelines provided by TED, including:

- The maximum subtitle reading speed is 21 characters per second;
- lines cannot exceed 42 characters, including white spaces;
- Subtitles cannot exceed 2 lines.

Participants were expected to use only the audio track from the provided videos (dev and test sets), the video track was of low quality and primarily meant to verify time synchronicity and other aspects of displaying subtitles on screen. That being said, the exploitation of the video was permitted.

The subtitling sub-track required participants to automatically subtitle audio-visual documents in German and/or Spanish, where the spoken language is always English. These documents were collected, similarly to last year, from the following sources:

- TED talks;[13]
- Physical training videos offered by Peloton;[14]
- TV series from ITV Studios.[15]

**Subtitle Compression.** The objective of the subtitle compression sub-track was to engage teams interested in the subtitling task but unable to build a complete automatic subtitling system. Participants were provided with automatic subtitles (in German and Spanish) generated by a non-participating system, namely the system presented in (Papi et al., 2023a), and asked to rephrase those that exceeded the reading speed constraint (more than 21 characters per second) to make them compliant. Time boundaries were to remain unchanged: only the text within a given time span had to be compressed when necessary. The original audiovisual documents (from the ITV test24 set of the subtitling sub-track) were also provided.

Although the subtitle compression task may appear simpler than subtitling, and it certainly is from the point of view of architectural complexity, it still presents its own difficulties. These challenges include those inherent in *text summarization*, such as identifying the main content of the original text, which must be preserved, and distinguishing accessory information, which can be omitted if necessary. Additionally, a peculiar challenge is that the text that needs to be reformulated is potentially error-prone and often does not consist of well-formed sentences but rather spans of text representing portions of sentences or words spanning contiguous phrases. It is expected that the most effective solutions are those capable of looking at the context, in an attempt to recover as much as possible the missing information in the text being processed.

### 4.2 Data and Metrics

#### 4.2.1 Automatic subtitling

**Data.** This sub-track proposed two training data conditions:

- **Constrained**: the official training data condition, in which the allowed training data is limited to a medium-sized framework[16] to keep the training time and resource requirements manageable;
- **Unconstrained**: a setup without data restrictions (any resource, pre-trained language models included, can be used) to allow also the participation of teams equipped with high computational power and effective in-house solutions built on additional resources.

| domain | set | AV docs | hh::mm | ref subtitles | |
|---|---|---|---|---|---|
| | | | | de | es |
| TED | dev | 17 | 04:11 | 4906 | 4964 |
| | test23 | 14 | 01:22 | 1375 | 1422 |
| | test24 | 16 | 01:50 | 1832 | 1826 |
| Peloton | dev | 9 | 03:59 | 4508 | 4037 |
| | test23 | 8 | 02:43 | 2700 | 2661 |
| | test24 | 4 | 01:40 | 1418 | 1574 |
| ITV | dev | 7 | 06:01 | 4489 | 4762 |
| | test23 | 7 | 05:08 | 4806 | 4896 |
| | test24 | 7 | 05:54 | 4564 | 4528 |

Table 5: Statistics of the dev and evaluation sets for the subtitling task.

For each language and domain, a development set and two test sets were released, that of the 2023 evaluation (**tst2023**), used for measuring progress over years, and a new one (**tst2024**). Table 5 provides some statistics on these sets.

**Metrics.** The evaluation was carried out from three perspectives, subtitle quality, translation quality, and subtitle compliance, through the following automatic measures:

- Subtitle quality vs. reference subtitles:
  - **SubER**, primary metric, used also for ranking (Wilken et al., 2022);[17]
- Translation quality vs. reference translations:
  - **BLEU**[18] and **CHRF**[19] via sacreBLEU;
  - **BLUERT** (Sellam et al., 2020).

Automatic subtitles are realigned to the reference subtitles using mwerSegmenter (Matusov et al., 2005)[20] before running sacreBLEU and BLEURT.

---
[13] https://www.ted.com/
[14] https://www.onepeloton.com
[15] https://www.itvstudios.com
[16] https://iwslt.org/2024/subtitling#training-data-allowed-for-constrained-conditions
[17] https://github.com/apptek/SubER
[18] sacreBLEU signature: nrefs:1|case:mixed|eff:no|tok:13a|smooth:exp|version:2.0.0
[19] sacreBLEU signature: nrefs:1|case:mixed|eff:yes|nc:6|nw:0|space:no|version:2.0.0
[20] https://www-i6.informatik.rwth-aachen.de/web/Software/mwerSegmenter.tar.gz

- Subtitle compliance:[21]
  - rate of subtitles with more than 21 characters per second (**CPS**);
  - rate of lines longer than 42 characters, white spaces included (**CPL**);
  - rate of subtitles with more than 2 lines (**LPB**).

#### 4.2.2 Subtitle compression

**Data.** No specific training data was released for this sub-track. Any solution was allowed, without limitations on the training data, including the use of LLM prompted for text compression (e.g. chatGPT). The original audio, though potentially helpful, could either be used or not by participants; its transcription with external tools (e.g. Whisper) was also permitted.

As a development set, a minimal example taken from the EuroParl Interviews benchmark (Papi et al., 2023a)[22] was released, where the non-participating subtitling system introduced in (Papi et al., 2023a)[23] was employed to generate automatic, sometimes non-compliant subtitles, which were associated with corresponding compliant reference subtitles.

The test set consists of German and Spanish automatic subtitles for the audiovisual documents defining the ITV test24 set of the subtitling sub-track; the same non-participating subtitling system was employed to generate the subtitles to be corrected.

**Metrics.** Since the text in subtitles has to be compressed to fulfill the CPS requirement, but at the same time its meaning should be preserved as best as possible, both **CPS** and **BLEURT** are considered primary metrics in the evaluation of compression quality.

### 4.3 Submissions

#### 4.3.1 Automatic subtitling

The subtitling sub-track saw the participation of three teams: APPTEK, the MT unit of Fondazione Bruno Kessler (FBK) with two different systems, and Huawei Translation Service Center (HW-TSC). The details about the participants' systems are provided below:

---
[21] https://github.com/hlt-mt/FBK-fairseq/blob/master/examples/speech_to_text/scripts/subtitle_compliance.py
[22] https://mt.fbk.eu/europarl-interviews/
[23] https://github.com/hlt-mt/FBK-fairseq/blob/master/fbk_works/DIRECT_SUBTITLING.md

**AppTek**: the cascade-based subtitling system developed by APPTEK[24] leveraging their in-production automatic captioning and translation offerings. A pipeline of in-house hybrid ASR, punctuation and inverse text normalization models is used to create English captions, which are segmented into blocks and lines via a neural segmentation model in combination with hard subtitling constraints, similar to Matusov et al. (2019). Time stamps follow from the HMM alignment of the first and last word in a block. In a second step, the generated source template is translated with customized transformer-based NMT models, for which full sentences are extracted and translations are reinserted into the template using a variant of the source-side segmentation method that enforces splitting into the existing blocks. The NMT models make use of preceding sentence context, and prefix tokens are used to provide genre and formality information (e.g. "talks" + "formal" for TED) and to control the length of the translation (Matusov et al., 2020). For the primary submission, the MT component is fine-tuned on high quality media and entertainment customer data. In addition, the following newly developed features are employed: automatic MT length token selection to condense translation only where necessary due to space constraints; extension of subtitle timings for lower reading speed; improved Spanish MT model. The contrastive submissions do not use these upcoming features. The second contrastive submission is created using APPTEK's general domain MT models, which are trained on publicly available data.

**FBK-AI4C$_{DIR}$** (Gaido et al., 2024a): the FBK's direct subtitling system is based on the transcription-free novel architecture, SBAAM or Speech Block Attention Area Maximization, introduced in (Gaido et al., 2024b). SBAAM leverages cross-attention scores to retrieve the timestamp information and is the first fully direct solution capable of producing automatic subtitles by eliminating any dependence on intermediate transcripts. It is the only system trained under constrained conditions, utilizing only the limited data provided by the IWSLT 2024 organizers. This includes non-subtitle material, which was automatically segmented into subtitles using the multimodal segmenter by Papi et al. (2022b). SBAAM is also employed as a reference system in the AI4Culture EU project[25] and is available at: https://github.com/hlt-mt/FBK-fairseq/blob/master/fbk_works/SBAAM.md.

---
[24] https://www.apptek.com/

**FBK-AI4C**$_{CSC}$ (Gaido et al., 2024a): the FBK's cascade subtitling system, developed by FBK within the AI4Culture project, exploiting pre-trained language models and, therefore, participating under the unconstrained conditions. The system is a cascade solution with Whisper (Radford et al., 2023) as the ASR model, and Helsinki Opus-MT (Tiedemann and Thottingal, 2020) as the MT model, together with additional components developed in-house. The cascade solution is publicly available at: https://github.com/hlt-mt/FBK-subtitler.

**HW-TSC** (Xie et al., 2024): the unconstrained cascade solution developed by HW-TSC, which relies on Whisper (Radford et al., 2023) to estimate both transcripts and word-level timestamps, on Bert-restore-punctuation[26] for retrieving punctuation and sentence segmentation, and on wav2vec2-large-960h-lv60[27] for the CTC-based force alignment between transcripts and translations, obtained by in-house MT models. The MT models (English to German and English to Spanish) were directly employed on the sentence-level ASR transcripts while the timestamps were left unchanged between transcripts and translations. Moreover, they are the only models among all participants that were specifically adapted to the domains of the audiovisual documents through ad-hoc domain adaptation.

### 4.3.2 Subtitle compression

Three teams participated in the sub-track: the FBK MT unit, the Huawei Translation Service Center (HW-TSC), and the Research Institute for Artificial Intelligence *Mihai Drăgănescu*, Romanian Academy (RACAI). The solutions they proposed differ from each other, although they share the use of Large Language Models as a common trait. Specifically:

**FBK** (Gaido et al., 2024a): the primary submission exploited GPT-4 (Achiam et al., 2023), which was prompted in zero-shot mode with an instruction asking the model to shorten the input text using the maximum number of characters compatible with the subtitle duration (value computed offline and passed as a parameter) while preserving the original words as much as possible. In the two contrastive runs, non-compliant subtitles were compressed by deleting function words from lists of different lengths.

**HW-TSC** (Xie et al., 2024): the subtitle compression method for the primary run is based on MT models, which are first employed for back-translating the non-compliant subtitles into English, and then to re-translate English into the original language (either German or Spanish) by setting a large beam size and a high length penalty, so that short translations are generated and rewarded. The still non-compliant subtitles are rewritten using the LLM Llama2 (Touvron et al., 2023), instructed with few-shot prompts to condense the input text. The two contrastive runs are variants of the primary one: in the first, the LLM is not applied and the compression is carried out only by the translation models; in the second, the subtitles of the primary run rewritten by either the MT model or the LLM which are still non-compliant are replaced by the original text.

**RACAI** (Gasan and Păiș, 2024): the submission involves generating multiple alternatives for the original non-compliant subtitle and selecting the one that maximizes both reading speed compliance (measured by CPS), and content similarity with the original subtitle (measured by ROUGE (Lin, 2004)). The alternatives are generated by i) rephrasing the subtitles using LLMs, specifically T5 (Raffel et al., 2020) and BART (Lewis et al., 2020), which were fine-tuned for the text summarization task, and ii) generating new subtitles through the automatic transcription of the original English audio using Whisper, translating them with NLLB (Costa-jussà et al., 2022), and then applying the LLMs as in the first method.

## 4.4 Results

The performance of runs for the two sub-tracks is presented and discussed separately in the following two subsections.

### 4.4.1 Automatic subtitling

Scores on tst2024 of all runs calculated using automatic metrics are shown in Tables 34 and 35, while Tables 37 and 38 refer to tst2023, where cumulative scores of runs submitted to the 2023 edi-

---
[25] https://pro.europeana.eu/project/ai4culture-an-ai-platform-for-the-cultural-heritage-data-space
[26] Bert-restore-punctuation1
[27] https://huggingface.co/felflare/bert-restore-punctuation

tion are also reported to allow the quantification of progresses.[28]

This year, unlike in the last edition, only one team (**FBK-AI4C**$_{\text{DIR}}$) participated with a system trained under constrained data conditions. Consequently, comparing its results with those of other participants is inherently unfair, and must be acknowledged if any comparisons are made. Notably, **FBK-AI4C**$_{\text{DIR}}$ is also the only direct system in the competition, highlighting that, despite advancements in direct approaches to spoken language processing, constructing cascade subtitling systems remains prevalent.

**tst2024**: Looking at performance in both German and Spanish, APPTEK achieved the best compromise between translation quality and subtitle compliance, as attested by the SubER values. It is interesting to note that their primary and contrastive1 systems provide better subtitle quality than contrastive2, especially on Spanish; since the first two systems featured fine-tuning on proprietary data, it can be hypothesized that such data is somehow "close" to the domains proposed in this evaluation campaign and therefore that the adaptation has rewarded these models. Overall, the new APPTEK systems (primary and contrastive1) surpass the one currently in production (contrastive2), although surprisingly the latter shows the best global SubER on German.

Focusing on the quality of the translation, in particular in terms of BLEURT, which better correlates with humans compared to BLEU and ChrF, the performance of HW-TSC's system is superior, likely because it is the only system explicitly fine-tuned on in-domain data. However, this system has not been optimized in terms of compliance, resulting in the lowest CPL score and, consequently, in high SubER scores.

The FBK cascade system, based mainly on pre-trained general-purpose models, shows high translation quality, especially in Spanish, and an acceptable conformity of subtitles. This proves the feasibility of building effective subtitling systems by appropriately assembling off-the-shelf models.

The FBK direct system, the only one based on a direct architecture and trained in constrained conditions, generated German subtitles with a surprisingly competitive overall SubER, despite the quality of the translation of the ITV and Peloton documents being lower compared to other systems. The good SubER probably derives from the ability of this system to satisfy subtitle compliance, which demonstrates the potential of the innovative approach it is based on. On the other hand, the gap in terms of translation quality on the two more challenging domains is in line with what already happened last year and with expectations, since unconstrained training allows building models on data more representative of real-life content.

**tst2023**: On German, the best systems are those by APPTEK which however did not improve the SubER score of the last year; in fact, there is an improvement in the quality of the translation which is counterbalanced by a worst CPS. Moreover, we note that the CPS of 4 out of 5 submissions from last year is better than any 2024 primary submission.

On Spanish, the improvements in the quality of the translations and of the SubER scores are generalized, while the CPS values worsen.

The progress made by the FBK team over the past year with their direct approach is notable in various aspects and for both languages, demonstrating the potential of end-to-end solutions for automatic subtitling.

#### 4.4.2 Human evaluation

This year's edition of the automatic subtitling sub-track introduces the human evaluation of the primary submissions for **tst2024** en→de. Table 24 shows the direct assessment scores obtained on a sample of 1000 subtitles randomly selected from the whole test set. The ranking differs from the automatic one based on SubER, particularly for the HW-TSC system which achieves the best DA value but the worst SubER score. This can be explained by the design of the human evaluation, which was focused on assessing the translation quality while segmentation and subtitle compliance were not directly considered. In fact, the human ranking closely agrees with the pure translation quality metrics, in particular BLEURT (see Table 24 vs. column Bleurt of Table 34). While this reassures the validity of using automatic MT metrics also for the domain of subtitle translation, in future evaluations we see the need to provide the evaluators with subtitles instead of plain text sentences so that subtitle compliance, segmentation and timing errors can be accounted for.

---
[28] In 2023, the evaluation was done on the three domains still proposed here plus one additional domain, EPTV; for the sake of comparability, in the computation of the cumulative scores of the 2023 runs, EPTV has been excluded.

### 4.4.3 Subtitle compression

Table 36 shows the results of the submissions to the subtitle compression sub-track in terms of BLEURT, computed against the reference subtitles and in charge of quantifying the translation quality, and CPS, as a measure of reading speed compliance. For the sake of discussion, the table also includes the results of a simple `Baseline` (id=[1]) and those of the provided subtitles to compress (id=[0]). In the baseline method, the original subtitles with a non-compliant reading speed were cut at the maximum number of characters compatible with the subtitle duration and without regard to maintaining the integrity of the words, which therefore may be incomplete.

The results indicate that the participants designed methods aimed to find a trade-off between translation quality and CPS compliance, standing the working point of their systems in the area between the two extremes represented by subtitles [0] and [1], which is highlighted in Figure 2.

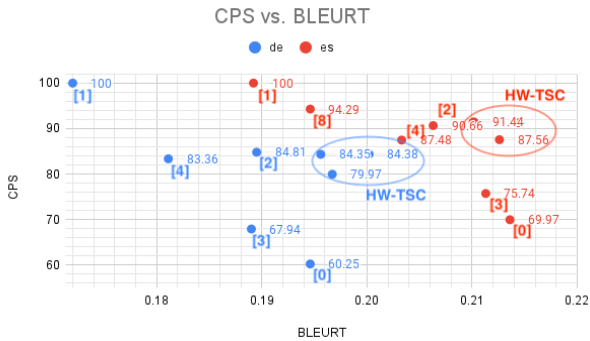

Figure 2: Scatter plot of compression results from Table 36.

Between [0] and [1], the subtitles generated by the contrastive **FBK** ([3,4]) and by the **RACAI** ([8]) systems are placed according to a nearly linear relationship. **HW-TSC**'s and, at a lesser extent, primary **FBK** ([2]) submissions differ markedly from this trend, thus demonstrating that it is possible to obtain a better compromise between the two contrasting features. In particular, the family of **HW-TSC** solutions is the most effective, approaching (in Spanish) or even overcoming (in German) the translation quality of the original subtitles, while achieving compliance for even more than 90% of the original subtitles. However, the noteworthy result of the **FBK** primary run shows the potential of prompting a generative LLM (GPT-4) to shorten subtitles; considering that it was done in zero-shot modality, there should be room for further improvements.

### 4.5 Conclusions

Overall, the second edition of the subtitling track continues to highlight the challenges and particularities of the automatic subtitling task. As in the previous edition, a clear gap in subtitle quality can be observed between the well-recorded, single-speaker, mostly formal style TED talk content that has traditionally been used for SLT evaluation at IWSLT, as opposed to the variety of audio conditions, dialog settings, language styles and speaking rates encountered in other types of content such as TV shows and sport videos. While no clear advancement in terms of best achieved translation quality or subtitle compliance compared to last year can be reported, remarkable improvements were achieved in the direct approach, which due to access to audio information during translation such as prosody, speaker changes and even speaker age/gender seems especially promising for subtitling of dialogs. The aspect of high speaking rates and the resulting necessity to condense subtitles down to a comfortable reading speed has been addressed and analyzed in isolation by the introduction of the subtitle compression task. Here, using LLMs for rephrasing has emerged as one of the promising approaches which was used by all participants.

## 5 Speech-to-Speech Translation

Speech-to-speech translation (S2ST) is a highly complex process involving the conversion of audio signals from one language to another. In offline translation, the system assumes that the entire audio is available before the translation process begins. This approach allows the translation system to process the audio input as a whole, enabling more effective speech recognition, semantic comprehension, and translation.

The main objective of this task is to encourage the development of automated methods for speech-to-speech translation that can perform efficiently and accurately in offline settings. Achieving this goal will not only advance the field but also contribute to improving access to information and communication across different languages and cultures.

## 5.1 Challenge

Participants built speech-to-speech translation systems from English into Chinese using any possible method, for example with a cascade system (ASR + MT + TTS or end-to-end speech-to-text translation + TTS) or an end-to-end or direct speech-to-speech system. Participants can use any techniques to boost the system performance.

## 5.2 Data and Metrics

**Data.** This task allowed the same training data from the Offline task on English-Chinese speech-to-text translation. More details are available in Sec. 2.2. In addition to the Offline task data, the following training data was allowed to help build English-Chinese speech-to-speech models and Chinese text-to-speech systems:

- **GigaS2S**, target synthetic speech for the Chinese target text of GigaST (Ye et al., 2023) that was generated with an in-house single-speaker TTS system;

- **aishell 3** (Shi et al., 2020), a multi-speaker Chinese TTS dataset.

**Metrics.** Since there was only one participant this year, we only conducted automatic evaluation in order to save resources.

**Automatic metrics.** To automatically evaluate translation quality, the speech output was automatically transcribed with a Chinese ASR system[29] (Yao et al., 2021), and then **BLEU**[30] (Papineni et al., 2002b), **chrF**[31] (Popović, 2015), and **COMET**[32] (Rei et al., 2022) were computed between the generated transcript and the human-produced text reference. BLEU and chrF were computed using SacreBLEU (Post, 2018).

## 5.3 Submissions

We only received submissions from one participant this year.

- HW-TSC (Wu et al., 2024) submitted three cascaded systems corresponding to three scenarios: constrained, constrained with large language models, unconstrained. All three scenarios employ a cascaded system that consists of an Automatic Speech Recognition (ASR) model, a translation model, and a Text-to-Speech (TTS) model. In the constrained scenario, the ASR model is trained on WeNet using constrained data. The translation model is a Transformer model trained using constrained data, with data enhancement, data denoising, and domain adaptation strategies applied, followed by model ensemble. The TTS model uses the VITS architecture. In the LLM constrained scenario, the ASR model is the same as in the constrained scenario. The translation model uses multiple LLMs for model ensemble, which are fine-tuned on llama2-13b using different strategies. The TTS model is the same as above. In the unconstrained scenario, the ASR model uses Whisper. The translation model employs multiple NMT models and LLMs for model ensemble. The TTS model remains the same as in the previous scenarios.

## 5.4 Results

Results by automatic metrics are shown in Table 39 in the Appendix.

## 6 Low-resource SLT

The 4[th] edition of the Low-resource Spoken Language Translation track focused on the translation of speech from a variety of data-scarce languages. The target language is typically a higher-resource one, generally of similar geographical or historical linkages. The goal of this shared task is to benchmark and promote speech translation technology for a diverse range of dialects and low-resource languages. While significant research progress has been demonstrated recently, many of the world's languages and dialects lack the parallel data at scale needed for standard supervised learning.

## 6.1 Challenge

This year's task significantly expanded the typological and geogrpahical diversity of the languages, language families, and scripts represented. The eight subtasks were:

- Bhojpuri → Hindi

- Marathi → Hindi

- Irish → English

---

[29] https://github.com/wenet-e2e/wenet/blob/main/docs/pretrained_models.en.md
[30] sacreBLEU signature: nrefs:1|case:mixed|eff:no|tok:zh|smooth:exp|version:2.3.1
[31] sacreBLEU signature: nrefs:1|case:mixed|eff:yes|nc:6|nw:0|space:no|version:2.3.1
[32] https://huggingface.co/Unbabel/wmt22-comet-da

- Maltese → English
- Bemba → English
- North Levantine Arabic → English
- Tamasheq → French
- Quechua → Spanish

Teams were allowed to submit to as few as one language pair, up to all eight. Both constrained and unconstrained submissions were allowed, to be separately ranked. For the constrained scenario, teams were only allowed to submit systems using the data provided by the shared task. For the unconstrained systems, teams were allowed to use any data as well as any pre-trained models.

### 6.2 Data and Metrics

Table 6 provides a summary of the training data that were part of the shared task. We describe in more detail the data for each language pair below.

**North Levantine Arabic–English (apc-eng)** Levantine Arabic, a well-established unit within the Arabic dialectal continuum, can be divided into at least three regional variants (Al-Wer and de Jong, 2017). *North* Levantine Arabic (also known as Syrian or Shami, ISO code: `apc`) is based on the urban speech of mainly Beirut and Damascus and is perceived as a separate linguistic unit (Ghobain, 2017).

Participants were provided with the UFAL Parallel Corpus of North Levantine 1.0 (Sellat et al., 2023), which includes about 120k lines of multi-parallel North Levantine-Modern Standard Arabic-English textual data, that can be downloaded from the LINDAT/CLARIAH-CZ Repository.[33] For additional speech data in North Levantine Arabic, participants were pointed to two LDC resources: the BBN/AUB DARPA Babylon Levantine corpus (Makhoul et al., 2005) and the Levantine Arabic QT Training Data Set 5 corpus (Maamouri et al., 2006).

Participants were also encouraged to use the Tunisian Arabic training data used in the last two years' shared task (LDC2022E01). This three-way parallel data corresponds to 160 hours and 200k lines of aligned audio in Tunisian speech, Tunisian transcripts, and English translations. Additionally, a number of OpenSLR resources in Modern Standard Arabic were highlighted: Tunisian Modern Standard Arabic speech and transcriptions[34], the MADCAT Arabic LDC corpus (Lee et al., 2012), the Arabic portion of theMediaSpeech corpus (Kolobov et al., 2021), and the Arabic speech to text Quran data.[35]

Overall, the provided resources were supposed to help participants, but only the unconstrained scenario was considered within this year's initial run of the apc-eng language pair.

The development[36] and test[37] data consist of recordings of native speakers of the dialect and is a mix of spontaneous monologues and dialogues on the topics of everyday life (health, education, family life, sports, culture), living abroad, and everyday life in Syria. The transcription and translation team consisted of students of Arabic at Charles University, with an additional quality check provided by the native speakers of the dialect.

**Bemba–English (bem-eng)** Bemba (also known as IciBemba) is a Bantu language (ISO code: `bem`), spoken predominantly in Zambia and other parts of Africa by over 10 million people. It is the most populous indigenous language spoken by over 30% of the population in Zambia where English is the lingua franca and official high-resourced language of communication. Bemba is native to the people of Northen, Luapula and Muchinga provinces of Zambia but also spoken in other parts of the country including urban areas such as Copperbelt, Central and Lusaka provinces by over 50% of the population (ZamStats, 2012).

The provided Bemba-English corpus (Sikasote et al., 2023a) consists of over 180 hours of Bemba audio data, along with transcriptions and translations in English. The dataset is comprised of recorded multi-turn dialogues between native Bemba speakers grounded on images.

In addition, we provided transcribed (28 hours) and untranscribed (60 hours) monolingual Bemba speech from Zambezi Voice (Sikasote et al., 2023b) and BembaSpeech (Sikasote and Anastasopoulos, 2022) datasets.

**Bhojpuri–Hindi (bho-hin)** Bhojpuri (ISO code: `bho`) belongs to the Indo-Aryan language group. It is dominantly spoken in India's western part of Bihar, the north-western part of Jharkhand, and the Purvanchal region of Uttar Pradesh. As per the 2011 Census of India, it has around 50.58 million speakers (Ojha and Zeman, 2020). Bhojpuri is spoken not just in India but also in other countries such as Nepal, Trinidad, Mauritius, Guyana, Suriname, and Fiji. Since Bhojpuri was considered a dialect of Hindi for a long time, it did not attract much attention from linguists and hence remains among the many lesser-known and less-resourced languages of India.

---
[33] http://hdl.handle.net/11234/1-5033
[34] https://www.openslr.org/46/
[35] https://www.openslr.org/132/
[36] http://hdl.handle.net/11234/1-5518
[37] http://hdl.handle.net/11234/1-5519

The provided Bhojpuri–Hindi corpus consists of 22.77 hours of Bhojpuri speech data (see Table 6) from the news domain, extracted from News On Air[38] and translated into Hindi texts.[39] Additionally, the participants were directed that they may use monolingual Bhojpuri audio data (with transcription) from ULCA-asr-dataset-corpus[40] as well as Bhojpuri Language Technological Resources (BHLTR) (Ojha et al., 2020; Ojha, 2019)[41] and Bhojpuri-wav2vec2 based model.[42]

**Irish–English (gle-eng)** Irish (also known as Gaeilge; ISO code: `gle`) has around 170,000 L1 speakers and 1.85 million people (37% of the population) across the island (of Ireland) claim to be at least somewhat proficient with the language. In the Republic of Ireland, it is the national and first official language. It is also one of the official languages of the European Union (EU) and a recognized minority language in Northern Ireland with the ISO `ga` code.

The provided Irish audio data were compiled from the news domain, Common Voice (Ardila et al., 2020),[43] and Living-Audio-Dataset.[44] The Irish–English corpus consists of 12 hours of Irish speech data (see Table 6), translated into English texts.

**Maltese–English (mlt-eng)** Maltese (ISO code: `mlt`) is a Semitic language, with a heavy influence from Italian and English. It is spoken mostly in Malta, but also in migrant communities abroad, most notably in Australia and parts of America and Canada.

The data release for this shared task consists of over 14 hours (split into dev and train) of audio data, together with their transcription in Maltese and translation into English. Participants were also allowed to use additional Maltese data including the text corpus used to train BERTu (Micallef et al., 2022), a Maltese BERT model, the MASRI Data speech recognition data (Hernandez Mena et al., 2020), and any data available at the Maltese Language Resource Server.[45]

**Marathi–Hindi (mar-hin)** Marathi (ISO code: `mar`) is an Indo-Aryan language and is dominantly spoken in the state of Maharashtra in India. It is one of the 22 scheduled languages of India and the official language of Maharashtra and Goa. As per the 2011 Census of India, it has around 83 million speakers which covers 6.86% of the country's total population.[46] Marathi is the third most spoken language in India.

The provided Marathi–Hindi corpus consists of 24.58 hours of Marathi speech data (see Table 6) from the news domain, extracted from News On Air[47] and translated into Hindi texts.[48] The dataset was manually segmented and translated by Panlingua.[49] Additionally, the participants were directed that they may use monolingual Marathi audio data (with transcription) from Common Voice (Ardila et al., 2020),[50] as well as the corpus provided by He et al. (2020)[51] and the Indian Language Corpora (Abraham et al., 2020).[52]

**Quechua–Spanish (que-spa)** Quechua (macrolaguage ISO code: `que`) is an indigenous language spoken by more than 8 million people in South America. It is mainly spoken in Peru, Ecuador, and Bolivia where the official high-resource language is Spanish. It is a highly inflective language based on its suffixes which agglutinate and are found to be similar to other languages like Finnish. The average number of morphemes per word (synthesis) is about two times larger than in English. English typically has around 1.5 morphemes per word and Quechua has about 3 morphemes per word.

---

[38] https://newsonair.gov.in
[39] https://github.com/panlingua/iwslt2024_bho-hi
[40] https://github.com/Open-Speech-EkStep/ULCA-asr-dataset-corpus
[41] https://github.com/shashwatup9k/bho-resources
[42] https://www.openslr.org/64/
[43] https://commonvoice.mozilla.org/en/datasets
[44] https://github.com/Idlak/Living-Audio-Dataset

[45] https://mlrs.research.um.edu.mt/
[46] https://censusindia.gov.in/nada/index.php/catalog/42561
[47] https://newsonair.gov.in
[48] https://github.com/panlingua/iwslt2023_mr-hi
[49] http://panlingua.co.in/
[50] https://commonvoice.mozilla.org/en/datasets
[51] https://www.openslr.org/64/
[52] https://www.cse.iitb.ac.in/~pjyothi/indiccorpora/

| Language Pairs | | Train Set | Dev Set | Test Set | Additional Data |
|---|---|---|---|---|---|
| Bhojpuri–Hindi | bho–hi | 19.88 | 2.07 | 0.82 | Monolingual audio with transcription (ASR) and monolingual text |
| Irish–English | ga–eng | 9.46 | 1.03 | 0.69 | IWSLT 2023 test set (with references) and MT data (monolingual and parallel corpora) |
| Marathi–Hindi | mr–hi | 15.88 | 3.66 | 0.61 | Monolingual audio with transcriptions (ASR), IWSLT 2023 test set (with references) and monolingual text |
| Maltese–English | mlt–eng | 10 | 2 | 2 | Monolingual audio with transcriptions (ASR), monolingual text |
| North Levantine–English | apc–eng | - | 2.5 | 1.85 | - |
| Tamasheq–French | tmh–fra | 17 | - | - | Untranscribed audio, data in other regional languages |
| Quechua–Spanish | que–spa | 1.60 | 1.03 | 1.03 | 48 hours of monolingual audio with transcriptions (ASR) and MT data (not transcribed) |
| Bemba–English | bem–eng | 167.17 | 5.89 | 5.83 | 28.12 hours of monolingual audio with transcriptions (ASR) and 60 hours of untranscribed audio data. |

Table 6: Training, development and test data details (in hours) for the language pairs of the low-resource shared task.

There are two main regional divisions of Quechua known as Quechua I and Quechua II. This data set consists of two main types of Quechua spoken in Ayacucho, Peru (Quechua Chanka ISO: `quy`) and Cusco, Peru (Quechua Collao ISO: `quz`) which are both part of Quechua II and, thus, considered a "southern" languages. We label the data set with `que` - the ISO norm for Quechua II mixtures.

The constrained setting allowed a Quechua-Spanish speech translation dataset along with the additional parallel (text-only) data for machine translation compiled from previous work (Ortega et al., 2020). The audio files for training, validation, and test purposes consisted of excerpts of the Siminchik corpus (Cardenas et al., 2018) that were translated by native Quechua speakers. For the unconstrained setting, participants were directed to another larger data set from the Siminchik corpus which consisted of 48 hours of fully transcribed Quechua audio (monolingual).

**Tamasheq–French** Tamasheq is a variety of Tuareg, a Berber macro-language spoken by nomadic tribes across North Africa in Algeria, Mali, Niger and Burkina Faso. It accounts for approximately 500,000 native speakers, being mostly spoken in Mali and Niger. This task is about translating spoken Tamasheq into written French. Almost 20 hours of spoken Tamasheq with French translation are freely provided by the organizers. A major challenge is that no Tamasheq transcription is provided, as Tamasheq is a traditionally oral language.

The provided corpus is a collection of radio recordings from Studio Kalangou[53] translated to French. It comprises 17 hours of clean speech in Tamasheq, translated into the French language. The organizers also provided a 19-hour version of this corpus, including 2 additional hours of data that was labeled by annotators as potentially noisy. Both versions of this dataset share the same validation and test sets. Boito et al. (2022) provides a thorough description of this dataset.

In addition to the 17 hours of Tamasheq audio data aligned to French translations, and in light of recent work in self-supervised models for speech processing, we also provide participants with unlabeled raw audio data in the Tamasheq language, as well as in other 4 languages spoken from Niger: French (116 hours), Fulfulde (114 hours), Hausa (105 hours), Tamasheq (234 hours) and Zarma (100 hours). All this data comes from the radio broadcastings of Studio Kalangou and Studio Tamani.[54]

---
[53] https://www.studiokalangou.org/

Note that this language pair is a continuation of last year's shared task, using the same test set as last year.

### 6.2.1 Metrics

We use standard lowercase BLEU with no punctuation to automatically score all submissions. Additional analyses for some language pairs are provided below. Were applicable, we also report chrF++ (Popović, 2015).

### 6.3 Submissions

The Shared Task received a record 69 submissions (for speech translation) from 12 teams for all 8 language pairs. The Shared Task also received 15 submissions for the speech recognition task of transcribing the input audio. They are described in detail below.

ALADAN (Kheder et al., 2024) provided a submission for the apc-eng direction, building upon a cascade of ASR and MT systems. The authors propose a character-level and word-level normalization process to handle the orthographic inconsistency between Arabic Dialects, merging words based on a combination of weighted Levenshtein distance and similarity of embeddings, as computed with a task-specific Word2vec model. Both ASR and MT systems are trained on a combination of public (e.g., IWSLT22 data, GALE speech corpus[55] for ASR, and, e.g., the UFAL parallel dataset provided by the organizers, Global Voices, LDC2012T09 for MT) and internal data (a combination of crowd-sourced and web-scrapped resources). For ASR, TDNN-F (Povey et al., 2018) and Zipformer (Yao et al., 2023) models are considered, that are firstly trained on a generic Arabic data, and then fine-tuned on a dialect-specific speech. For MT, both encoder-decoder models and instruction-following LLMs are explored. The primary solution uses both ASR systems combined with the ROVER (Fiscus, 1997) algorithm, with the MT step performed by the fine-tuned Command-R[56] LLM, enhanced by MBR decoding and checkpoint averaging. Contrastive submissions differ in the MT step, with the first one using the final checkpoint of the fine-tuned LLM, and the second one using a Transformer-based NLLB model.

BITSP (Anand et al., 2024) submitted systems for the Bhojpuri to Hindi and Marathi to Hindi tasks. Their approach relied on cascading transcriptions which were piped into translation systems. They used a fine-tuned Whisper model for Marathi-Hindi and an vakyansh-wav2vec model for Bhojpuri-Hindi (Chadha et al., 2022; Gupta et al., 2021). Translation was done using fine-tuned NLLB for both tasks (NLLB Team et al., 2022). They also looked at using sentence-embeddings generated using the MuRIL (Multilingual Representations for Indian Languages) (Khanuja et al., 2021) model for the Marathi-Hindi task.

HW-TSC (Jiawei et al., 2024) participated in the apc-eng direction with a cascade solution based on the off-the-shelf Whisper (Radford et al., 2022) model for ASR combined with a Transformer-based MT model trained from scratch for Arabic-to-English translation. The MT system (35 encoder layers, 3 decoder layers, with $d_{hidden}$ = 512 and $d_{FFN}$ = 2048) was trained on the mix of publicly available (e.g., OpenSubtites, GlobalVoices, TED) and in-house corpora, both filtered based on sentence embeddings extracted with LaBSE (Feng et al., 2022). No dialect-specific datasets were used for training directly. Instead, an in-domain model was fine-tuned on the validation set to score the training samples using domain features (Wang et al., 2020c), with the highest-scoring subset explored for the final fine-tuning.

JHU (Robinson et al., 2024) provided systems for all eight language pairs. The main effort of their work revolved around fine-tuning large and publicly available models in three proposed systems, one cascaded and two end-to-end. For the cascaded system, they proposed fine-tuning Whisper transcription (not translation) and then piping that output to a fine-tuned NLLB model. For the end-to-end systems, they fine-tuned for translation directly on SEAMLESS4MT v2 and Whisper translation (not transcription). In addition, they looked at a variety of different training paradigms such as intra-distillation (Xu et al., 2022), joint training, multi-task learning, curriculum learning, and pseudo-translation. The best-performing approach, similar to the broader results of this shared task differed for different language pairs. However, fine-tuned SEAMLESSM4T v2 tends to perform best for source languages on which it was pre-trained. Additionally, while multi-task training helps Whisper fine-tuning, in general cascaded systems with Whisper and NLLB tend to outperform Whisper alone. Finally, intra-distillation was shown to help NLLB fine-tuning.

---
[54]https://www.studiotamani.org/
[55]https://arabicspeech.org/resources
[56]https://cohere.com/command

| Team Name | Language Pairs | | | | | | | |
|---|---|---|---|---|---|---|---|---|
| | apc-eng | bem-eng | bho-hin | gle-eng | mlt-eng | mar-hin | que-spa | tmh-fra |
| SETU-DCU (Zafar et al., 2024) | | | | ✓ | ✓ | | | |
| UM (Nabhani et al., 2024) | ✓ | | | | ✓ | | | |
| UoM (Abela et al., 2024) | | | | | ✓ | | | |
| QUESPA (Ortega et al., 2024) | | | | | | | ✓ | |
| JHU (Robinson et al., 2024) | ✓ | ✓ | ✓ | ✓ | ✓ | ✓ | ✓ | ✓ |
| HW-TSC (Jiawei et al., 2024) | ✓ | | | | | | | |
| ALADAN (Kheder et al., 2024) | ✓ | | | | | | | |
| KIT (Li et al., 2024c) | ✓ | ✓ | | ✓ | ✓ | | | |
| BITSP (Anand et al., 2024) | | | ✓ | | | ✓ | | |
| YMOSLEM (Moslem, 2024) | | | | ✓ | | | | |
| UoM-DFKI (Rishu et al., 2024) | | | | | ✓ | | | |
| Total Teams per Lang Pair: | 5 | 2 | 4 | 3 | 5 | 2 | 2 | 1 |

Table 7: Breakdown of the teams and the language pairs subtasks that they participated in for the Low-Resource Shared Task.

KIT (Li et al., 2024c) participated in the Maltese-to-English, Bemba-to-English, North Levantine Arabic-to-English tasks in the unconstrained condition. They leveraged pretrained multilingual models by fine-tuning them for the target language pairs, looking at SeamlessM4T, NLLB (NLLB Team et al., 2022), and MMS (Pratap et al., 2024). Due to the large size of the models, they experimented with adapter fine-tuning to reduce the number of trainable parameters using LORA (Hu et al., 2021) and package PEFT (Mangrulkar et al., 2022). They were also able to show that Minimum Bayes Risk is effective in improving speech translation performance by combining systems in all of their language pairs.

SETU-DCU (Zafar et al., 2024) presented systems for two language pairs, Irish–English and Maltese–English. Both of their submissions, despite lower performance on the Irish (GA) task, were on the unconstrained condition configuration. There were two submissions to the Maltese (MLT) task ranging from 44.7 to 52.6 BLEU and one submission to the GA task at 0.6 BLEU.

The MLT results of 52.6 BLEU were favorable due to SETU-DCU's primary submission based on a cascaded (ASR to MT) setup of a Whisper (Radford et al., 2022) ASR system used in conjunction with an MT system based on the NLLB (NLLB Team et al., 2022) where both systems were fine-tuned on the Maltese–English data provided. Additionally, their cascaded Contrastive 1 system which used mBart-50 for decoding, scored 44.7 BLEU showing that the use of the NLLB system augmented performance by nearly 8 BLEU points. Further results can be attributed to data preparation such as removing unnecessary data chunks from the dataset, eliminating special characters, and converting the sentences to lowercase along with the following hyper-parameter configuration: batch size of 16, learning rate of 1e-5, 500 warmup steps, 30,000 max steps, per-device eval batch size of 8, generation max length of 225, and intervals of 1,000 steps for saving and evaluating, and 25 steps for logging.

SETU-DCU's submission for the unconstrained GA task performed poorly compared to other systems submitted. It consisted of a direct speech translation system using the Whisper *small* model by first resampling data at 16 khz and us-ing the following hyper-parameter configuration: batch size of 16, learning rate of 1e-5, 500 warmup steps, 1 gradient accumulation steps, generation max length of 225, and intervals of 500 steps for saving and evaluating. The model was fine-tuned over three epochs. Their only submission used Whisper for fine-tuning; however, their claim is that since the data Whisper was trained on did not contain GA at the time of fine-tuning, generation was inconsistent.

UoM-DFKI (Rishu et al., 2024) participated in the Maltese to English shared task using two popular end-to-end pretrained models, Whisper and wav2vec 2.0. They hypothesised that Maltese shares lots of vocabulary with Arabic and Italian and would therefore have good cross-lingual transfer ability due to pretraining data in those models. In addition, they investigated other popular neural models, BERT (Devlin et al., 2019) which they decided against making a formal submission, and mBART (Liu et al., 2020b) which was used as their contrastive submission. Overall, the end-to-end system performed much better than the contrastive submission.

UoM (Abela et al., 2024) participated in the constrained task of the Maltese to English translation language pair. Their approach relied on a cascaded system consisting of a pipeline containing: a DeepSpeech 1 ASR system (Hannun et al., 2014), a KenLM model to optimise the transcriptions (Heafield, 2011), and finally an LSTM machine translation model. For their ASR system, they trained using the MASRI dataset and CommonVoice and used a much smaller layer size (64) than normal due to the lack of large amounts of data. These outputs were then used to decode using a 3-gram statistical language model trained on Malti v4.0. The translation system was implemented using fairseq (Ott et al., 2019) comparing both transformer and LSTM architectures, with their best performing system using LSTMs. The authors hypothesize that this was due to the very small amount of data available as a bitext.

UM (Nabhani et al., 2024) competed in the unconstrained task for Maltese-English and North Levantine Arabic-English spoken language translation using a pipeline approach. For the ASR component of their systems, they relied on fine-tuning XLS-R using 50 hours of Maltese speech data. To correct outputs, they relied on the statistical toolkit KenLM (Heafield, 2011). Machine translation was then done using a fine-tuned version of the 1.3B parameter NLLB model (NLLB Team et al., 2022). They experimented with a variety of data sources such as CommonVoice, MASRI, and OPUS-100.

YMOSLEM (Moslem, 2024) The Yasmin Moslem team (independent researcher) presented an end-to-end approach for speech translation from spoken Irish to written English. Their models are based on Whisper, utilizing small, medium, and large versions. The primary system employs Whisper-large, which has been fine-tuned using the official training data, supplemented with synthetic audio data and the data augmentation technique involving white noise and voice activity detection.

The synthetic audio data was generated using Azure's text-to-speech service, applied to the Wikimedia dataset comprising 7,545 text segments. The resulting synthetic audio dataset consists of two parts: one featuring a female voice (OrlaNeural) and the other a male voice (ColmNeural). This resulted in a total of 15,090 utterances, with each text segment used to generate a synthetic speech segment for each voice. The same approach has been applied to 3,966 text segments coming from the SpokenWords dataset.

In addition to the official IWSLT-2023 training dataset and the aforementioned synthetic audio dataset, the Irish portion of the FLEURS dataset, the Bitesize dataset, and the SpokenWords dataset were utilized to fine-tune the Whisper-Large model. Note that the Irish portion of the Spoken Words dataset has been translated into English using the Google Translation API.

QUESPA (Ortega et al., 2024) submitted six total systems consisting of three *constrained* and three *unconstrained* systems. Team QUESPA were able to improve the previous year's results despite the data remaining the same as last year's ranging from 1.4 to 2.0 BLEU for the constrained task and 11.1 to 19.7 BLEU for the unconstrained one. This year QUESPA provided developmental results on several models that used mel-filter bank (MFB) features extracted using Fairseq (Wang et al., 2020a) were included that show the effect of the *s2t transformers* model type size ranging from extra-small to large.

QUESPA's *Constrained* systems did not vary

| Language Pair | Winning Team | System | Constrained? | BLEU |
|---|---|---|---|---|
| apc-eng | ALADAN | primary | no | 28.71 |
| bem-eng | JHU | primary | no | 32.60 |
| bho-hin | JHU | primary | no | 24.40 |
| gle-eng | JHU | contrastive1 | no | 16.00 |
| mlt-eng | KIT | primary | no | 58.90 |
| mar-hin | IITM | primary | no | 47.20 |
| que-spa | QUESPA | contrastive1 | no | 19.70 |
| tmh-fra | baseline | primary | no | 8.83 |

Table 8: Winning submissions for each language pair of the Low-Resource Shared Task.

much from last year's systems as far as system architecture is concerned. However, they were able to identify a caveat in the training data set which contains audio wav files of lengths from 1 to 30 seconds while the developmental and test sets were all of 30 seconds in length. Their opinion is that the varied length warranted a severe hyper-parameter empirical search resulting in a Primary system that scored 2.0 BLEU with the following configuration of a Fairseq (Wang et al., 2020a) speech translation model based on mel-Filter Bank features: extra-small transformer, 6 encoder layers, 3 decoder layers, Adam optimization, 500 epochs and a learning rate of .0002 while using an average of the last 10 checkpoints which outperformed the same model with other hyper-parameters from last year. Their Contrastive 1 system, similar to the primary system, introduced a new concept of data augmentation in combination with a medium transfomer (s2t_transformer), 12 encoder layers, 6 decoder layers, and 8 attention heads and 200 epochs. More importantly, in Contrastive 1 they introduced audio augmentation via LibRosa[57] where the translation was the same but four audio techniques were introduced: *Noise* (0.009 aggregation), *Roll* ($sr/10$), *Time*(0.4), and *Pitch* (-5) to create 4-fold sets of the original. Additionally, QUESPA's Contrastive 1 system removed SpecAugment as an audio augmentation technique. Finally, the Contrastive 2 system from Team QUESPA were identical to the primary system with the change of epochs to 400 and model type to a medium-size (s2t_transformer).

QUESPA's *Unconstrained* systems were a novel introduction for the QUE–SPA task and outperformed last year's best systems. Their primary system introduced the SpeechT5 (Ao et al., 2022)

ASR PLM which consists of 12 Transformer encoder blocks and 6 Transformer decoder blocks, with a model dimension of 768, an internal dimension (FFN) of 3,072, and 12 attention heads. It used normalized training text from the LibriSpeech language model as unlabeled data, which consisted of 400 million sentences and fine-tuned on the competition data while optimizing with Adam and a learning rate maximum of 0.0002. Fine-tuning was performed using the SpeechT5 fine-tuning recipe[58] for Speech-Translation with the same hyperparameter settings. Additionally, their primary system used a data augmentation technique (noise, distortion, duplication)[59] (Ma, 2019) for total of 120h: 60h original + 60h synthetic data scoring 16.0 BLEU, higher than previous year's results. For Contrastive 1, QUESPA introduced a combination of more data by manually translating Quechua to Spanish 55 hours of the total set along with an additional 19 minutes of Guarani and 29 minutes of Bribiri from the AmericasNLP[60] shared task. On top of that, they applied two data augmentation techniques: (1) *nlpaug* (Ma, 2019) and (2) DA-TTS (Zevallos et al., 2022), which involves generating synthetic text and audio using a de-lexicalization algorithm and a TTS system for the source language (Quechua). These two data augmentation techniques generated 62 hours and 50 hours respectively. Altogether, they used a total of 167h and 48 min: 55h (new dataset) + 48 min (ANLP dataset) + 62h nlpaug + 50h DA-TTS. The Contrastive 1 system was QUESPA's best system scoring 19.7 BLEU. The Contrastive 2 system was also newly introduced with the use of Whisper medium-size, multi-lingual model for ASR in a cascade approach basically replacing last year's "fleurs" ASR system. The MT system was identical to the one they used last year called FloresMT (Ortega et al., 2023). QUESPA's Contrastive 2 system resulted in a score of 11.1 BLEU.

---
[57] https://librosa.org/
[58] https://github.com/microsoft/SpeechT5/tree/main/SpeechT5
[59] https://github.com/makcedward/nlpaug
[60] https://turing.iimas.unam.mx/americasnlp/2022_st.html

### 6.4 Results

Table 8 summarizes the winning submissions for each language pair. Detailed results for all teams' systems and settings are available in Appendix B.5.

Of the 8 language pairs, 5 different teams had the top performing system on at least one language pair. This shows how competitive the shared task was, and that a multitude of approaches are helpful for low-resource speech translation. Additionally, no team was able to beat the baseline on the Tamasheq-French direction (which corresponds to last year's best system). This suggests that there continues to be lots of room for improvement and that this remains an active area of research.

Compared to previous iterations of the shared task, many of the language pairs had marked improvements with large gains in the official automatic metrics. For example, BLEU scores for Maltese-English and Marathi-Hindi are in the 40s and 50s. Furthermore, for North Levantine Arabic-English, Bemba-English, and Bhojpuri-Hindi are above 20 BLEU points. Even for Quechua-Spanish, the least resourced language pair, the best submission's BLEU score is almost 20 points.

This marks stark improvements from last year's shared task systems for some language pairs. In Marathi-Hindi, the best system in 2023 achieved a BLEU score of 39.7, with this year's best system improving by more than 7 BLEU points. Similarly, the improvements in the quality and quantity of the Maltese data lead to a more than 50 BLEU points improvement compared to last year. For Irish and Tamasheq, the performance increases are more modest, about 1 to 2 BLEU points in each, compared to the 2023 Shared Task.

For the language pairs included for the first time in the shared task, we find that Bemba-English and Bhojpuri-Hindi end up with decent systems, a result of high-quality data availability: for instance, Bemba-English has an order of magnitude more training data –167h– than any other language pair in our shared task); and Bhojpuri is the second most "high-resourced" language in our set, with almost 22 hours of speech translation data.

Within the systems submitted to the initial run of the North Levantine Arabic-English language pair, all of the primary submissions are based on a pipeline approach exploring ASR and MT, with a single submission combining E2E and cascaded systems. Since the popular NLLB model explored by several submissions supports an input/output combination of dialectical Arabic/English and a large-scale, parallel textual dataset of Levantine Arabic was provided, the participating teams mainly struggled with the ASR component. The winning submission by AL-ADAN, which outperformed a second-place team by over 8 BLEU points, uses an internal dataset of Levantine speech to boost the performance of their ASR component. While the data used for fine-tuning the MT system is comparable between the submissions, ALADAN explored a much larger, prompt-driven LLM compared to the 600M/1.3B NLLB variants explored by other teams.

We note that almost all submissions followed the unconstrained setting – a clear indication that pre-trained multilingual systems seem to be the best option for building ST for low-resource languages, at least under the current data, architectural, and compute constraints.

## 7 Automatic Dubbing

### 7.1 Challenge and Test Sets

Dubbing is a form of speech translation where the user can not only hear the translated speech, but also can often see the original speaker. This adds numerous challenges and constraints, including isochrony (does the new translation respect the timing of the original speech), phonetic synchrony or lip sync (is the new speech compatible with the mouth movements of the original speaker, if visible), kinesic synchrony (is the new speech consistent with visible body movements of the original speakers), and others (Mayoral et al., 1988; Chiaro, 2009; Chaume, 2020; Brannon et al., 2023).

For English→Chinese, we use the ITV test set from subtitle task. We manually selected 10min sections from each of clip 15, 16, 18, 19, and 21. The 10min sections were manually selected with several goals:

1. Speech is fairly clear

2. A mix of on-/off-screen dialogues
3. A diverse set of genders and accents
4. Avoid excessive profanity
5. Avoid opening/closing credits

German→English followed the same setup as the submissions from last year (Chronopoulou et al., 2023; Pal et al., 2023; Rao et al., 2023).

### 7.2 Submissions

This task received a total of four English→Chinese submissions (see Table 9): one end-to-end dubbing submission and three participants in the offline speech translation task (speech to text) scored our challenge set (set5). For the offline submissions, we utilized the provided translations to generate dubs.

We also received one submission (Li et al., 2024b) for German→English. We chose to focus on English→Chinese for evaluation due to the availability of the offline speech systems to compare against, which should represent strong speech translation models (but not dubbing specific models).

The process of generating dubs from text translations involved several steps. First, due to the absence of source language subtitles, we downloaded subtitles from an open-source website and manually time align the five clips. Each time aligned sentence was then split at commas and full stops to create manageable segments for processing, while keeping a track of original sentences and time-stamps.

Similarly, the translations from the three submissions were also split at commas and full stops. We used Vecalign (Thompson and Koehn, 2019, 2020) a tool for sequence alignment, in conjunction with LASER-2 embeddings (Heffernan et al., 2022), to align the source language with the target language. This ensured that the meaning and context of the translated text matched the original as closely as possible. Timestamps were then projected from the source to the target language, providing a temporal map for the dubbing process.

For each sentence, we employed Amazon Polly, a text-to-speech service, to generate the corresponding speech. We also used the duration of the source speech segment as a constraint to generate target speech with Polly. Polly allowed this by adding a flag with $max\_durations$, where the generated speech cannot go beyond maximum duration. We used Zhiyu standard voice as that allowed use of this flag via SSML wrapper. Adding duration constraint essentially ensured that the target speech did not exceed the length of the source speech. Typically, the target speech was shorter than the source speech, so we filled the remaining portion with silences to maintain synchronization.

We synchronized the start time of the target speech with the source speech using the previously obtained timestamps to ensure that the dialogue matched the visual cues accurately. Finally, we concatenated the target speech segments to form the complete clip.

### 7.3 Metrics and Results

We report speech overlap (between the original audio and the dubbed audio) in Table 10. For reference, in a large corpus of professionally dubbed media, human speech overlap between original and dubbed speech is about 0.658 (mean) and 0.731 (median) (Brannon et al., 2023). The dubbing submission HWTSC-Dubbing is similar to the human statistics, while the cascaded systems generated in part by the task organizers perform substantially worse.

We report PEAVS (Perceptual Evaluation of Audio-Visual Synchrony) score (Goncalves et al., 2024), an automatic metric with a 5-point scale that evaluates the quality of audio-visual synchronization, in Table 12. PEAVS is the only AV sync evaluation metric that is grounded in human judgements as it is trained on a large Audio-Visual synchrony benchmark for "in-the-wild" videos. In our case, we use PEAVS for evaluating the quality of synchrony in the generated dubs. As expected for a system optimized with speech timing in mind, HWTSC-Dubbing performs best here.

Table 12 also reports BLASER 2.0-QE scores. BLASER 2.0-QE is a reference-free modality-agnostic automatic metric for speech translation quality (Seamless Communication et al., 2023). It only supports short-form speech, so we segment the full speech into sentences as mentioned in Section 7.2 and report average scores. Surprisingly, the dubbing submission performs the best at this metric, even though it is optimized for both translation quality and timing. It is worth noting that the segments being evaluated are quite short, often much shorter than typical sentences in written text, and lack of domain context has been shown to be problematic in machine translation metrics even for normal length sentences (Läubli et al., 2018; Toral et al., 2018; Vernikos et al., 2022). BLASER 2.0-QE is not trained on dubbing data, so there is likely degradation due to domain mismatch (Zouhar et al., 2024).

| Submission | Submission Type |
| --- | --- |
| HWTSC-Dubbing (Li et al., 2024b) | Dubbing |
| HWTSC-Offline (Wu et al., 2024) | Offline Speech Translation Challenge Set |
| NYA-Offline (Zhang et al., 2024) | Offline Speech Translation Challenge Set |
| CMU-Offline (Yan et al., 2024) | Offline Speech Translation Challenge Set |

Table 9: Submissions to the Dubbing Track

We report two measures of lip sync, both from Prajwal et al. (2020): LSE-D (lip-sync error distance) and LSE-C (Lip Sync Error - Confidence) (see Table 13). LSE-D measures the accuracy of audio-visual synchronization by identifying the offset with the smallest distance between audio and video features. LSE-C measures the confidence in this synchronization by comparing the best match's distance to those of adjacent offsets, with higher values indicating greater confidence. In essence, LSE-D tells us how well the audio and video are synchronized, while LSE-C tells us how sure the model is about that synchronization. HWTSC-Dubbing performs the best at LSE-D on average, although one strange result is that the metric prefers HWTSC-Dubbing to the original audio in two of the test sets, which does not make sense. Another surprise is that CMU-Offline slightly outperforms HWTSC-Dubbing on the LSE-C metric.

We also conduct human judgements to evaluate translation quality and naturalness. We evaluate the first 20 sentences of each clip based on the rubric (Table 11), and report the average score for each submission in Table 12. In general, the dubbing system produces more natural speech but sometimes less accurate translation than the offline systems. The offline systems oftentimes have to speed up the speech synthesis to match the original duration of a sentence, leading to hard-to-recognize speeches.

## 8 Indic Languages Track

In the realm of spoken language processing, speech-to-text translation (ST) holds a crucial role at the intersection of natural language processing. The primary aim of ST is to convert spoken language from one linguistic context into written text in another language. This typically involves using Automatic Speech Recognition (ASR) to convert speech in the source language into text, followed by Machine Translation (MT) to translate the source language text into the target language. ST is a multimodal task that takes speech input and produces output in text format. Furthermore, it is inherently multilingual, taking speech input in one language and generating text output in another. Traditionally, human language translators proficient in both the source and target languages have handled this task. However, the scarcity of translators fluent in multiple languages has created a pressing need for a dedicated model tailored to excel in the unique realm of ST tasks across diverse languages. Recent advancements in ST have predominantly focused on high-resource languages, leaving a significant gap for low-resource languages that face a substantial catch-up journey. The attention imbalance is primarily due to the scarcity of data for low-resource languages, as most deep-learning models depend on data abundance. Acquiring such data for low-resource languages poses a formidable challenge.

While a considerable body of research is dedicated to ST across diverse language families, a noticeable gap exists in investigating this domain concerning low-resource Indian languages. Currently, there are no datasets specifically designed for the ST task in Indian languages, covering both the Indo-Aryan and the Dravidian language families. This research aims to create either an End-to-End (E2E) or a Cascaded ST model

This Indic track aims to establish an ST translation model that spans a diverse array of dialects and low-resource languages originating from the Indo-Aryan and Dravidian language families in India. Given that a significant portion of the data is sourced from very low-resource languages, these languages remain largely unexplored in the realm of speech translation. Compounding this challenge is the fact that many of the target languages are distantly related to English. Consequently, we anticipate that relying solely on pre-trained models may encounter numerous obstacles. The dataset provided will serve as the inaugural benchmark and gold standard dataset, encompassing all Indian languages. We aspire for participants to develop systems capable of real-world deployment in the future.

| Test Set | 15 | 16 | 18 | 19 | 21 | Average |
|---|---|---|---|---|---|---|
| HWTSC-Dubbing | 0.721 | 0.585 | 0.718 | 0.749 | 0.715 | 0.698 |
| HWTSC-Offline | 0.281 | 0.228 | 0.277 | 0.374 | 0.238 | 0.280 |
| NYA-Offline | 0.316 | 0.194 | 0.274 | 0.385 | 0.225 | 0.279 |
| CMU-Offline | 0.365 | 0.206 | 0.323 | 0.372 | 0.253 | 0.304 |

Table 10: Speech Overlap (↑), computed on speech segments as detected by silero-vad (Silero Team, 2021).

| Score | Description |
|---|---|
| 1 | Speech is not natural at all and/or the translation has nothing to do with the source. |
| 2 | Speech is not natural but you can understand why some of the words in the translation are there. |
| 3 | Speech is partially matching speakers lips and/or is a bit natural as well as the meaning of the source sentence are adequately transferred into the target language. |
| 4 | Speech naturalness is of acceptable quality and the meaning of the source sentence is mostly preserved. |
| 5 | Speech is mostly natural and the translation is almost perfect or is a good paraphrase of reference. |
| 6 | Speech looks completely natural and the translation is perfect in every sense of the word. |

Table 11: Dubbing human evaluation rubric.

| Model | PEAVS (↑) | BLASER-QE (↑) | Human Evaluation (↑) |
|---|---|---|---|
| Original | 3.82 ±0.41 | – | – |
| HWTSC-Dubbing | 3.05 ±0.45 | 3.25 | 3.9 |
| HWTSC-Offline | 1.33 ±0.37 | 3.07 | 3.5 |
| NYA-Offline | 1.28 ±0.31 | 3.03 | 3.3 |
| CMU-Offline | 1.28 ±0.31 | 3.07 | 3.2 |

Table 12: PEAVS (Perceptual Evaluation of Audio-Visual Synchrony) score (Goncalves et al., 2024), BLASER 2.0-Q, a reference-free modality-agnostic automatic metric for speech translation quality (Seamless Communication et al., 2023), and human evaluation results.

| Test Set | 15 | 16 | 18 | 19 | 21 | Average |
|---|---|---|---|---|---|---|
| Original | 8.220 | 7.258 | 11.553 | 9.311 | 10.197 | 9.308 |
| HWTSC-Dubbing | 11.969 | 5.398 | 11.341 | 11.887 | 11.200 | 10.359 |
| HWTSC-Offline | 13.596 | 12.219 | 12.024 | 12.748 | 8.437 | 11.805 |
| NYA-Offline | 14.094 | 11.539 | 10.488 | 12.833 | 8.409 | 11.473 |
| CMU-Offline | 14.793 | 12.834 | 12.499 | 12.817 | 7.933 | 12.175 |

Table 13: Lip sync error distance (LSE-D, ↓) (Prajwal et al., 2020) at clip level.

| Test Set | 15 | 16 | 18 | 19 | 21 | Average |
|---|---|---|---|---|---|---|
| Original | 3.714 | 0.656 | 1.190 | 3.340 | 1.443 | 2.069 |
| HWTSC-Dubbing | 0.638 | 1.011 | 1.463 | 1.185 | 0.893 | 1.038 |
| HWTSC-Offline | 0.477 | 0.834 | 1.095 | 1.355 | 0.849 | 0.922 |
| NYA-Offline | 0.674 | 0.567 | 1.153 | 0.944 | 0.697 | 0.807 |
| CMU-Offline | 0.706 | 0.971 | 2.143 | 1.019 | 0.718 | 1.112 |

Table 14: Lip-sync error confidence (LSE-C, ↑) (Prajwal et al., 2020) at clip level.

### 8.1 Challenge

The Indic shared task consists of ST for three language pairs from English (en) to Hindi (hi), Tamil (ta), and Bengali (bn). The ST data for all these three language pairs is derived from the IndicTEDST dataset (Sethiya et al., 2024). The submissions are allowed for both the constrained and the unconstrained cases. The constrained case involves only the data provided in the task. The unconstrained case can utilize either the data provided in the challenge or any external data, along with any pre-trained models. The submissions are also allowed for the cascade and end-to-end models for all the language pairs. Thus, the task accepts the following cases for all three language pairs (en-hi, en-ta, and en-bn):

- End-to-end + Constrained
- End-to-end + Unconstrained
- Cascade + Constrained
- Cascade + Unconstrained

### 8.2 Data and Metrics

The ST task data for the Indic track encompasses three Indian languages representing diverse language families. The languages included in this shared task are Hindi (hi), Bengali (bn), and Tamil (ta), originating from the Indo-Aryan and Dravidian language families. The dataset includes speech and text (transcriptions) in English (source language) and text (translations) in Hindi, Bengali, and Tamil (target languages).

The data for this Indic track comprises a ST corpus that includes 3 low-resource Indian languages. The data is curated from the TED talks with Indic translations, usually a talk spans from 3 minutes to 15 minutes. A segmentation of the audio files in the form of YAML is provided with the data. Table 15 illustrates the consistency maintained across all corpora, with an equal number of lines in their .en, .lang, and .yaml files. However, due to inherent linguistic differences, the number of tokens in the .en and .lang files varies. The count of audio files corresponds to the number of distinct talks, each delivered by an individual speaker. Additionally, the speech hours indicate the cumulative speech duration in a given language. Each parameter is meticulously categorized into test, train, and valid subsets, establishing a comprehensive and structured dataset.

| Lang en→ | Split | #Lines | #Tokens (en) | #Tokens (lang) | #Audio files | #Speech (hrs) |
|---|---|---|---|---|---|---|
| bn | test | 1.1 | 19.3 | 17.3 | 15 | 2.09 |
|  | train | 5.1 | 89.4 | 80.4 | 106 | 9.20 |
|  | valid | 1.3 | 22.1 | 20.4 | 30 | 2.30 |
| hi | test | 7.2 | 118.6 | 138.0 | 75 | 13.52 |
|  | train | 45.8 | 752.6 | 890.5 | 528 | 76.46 |
|  | valid | 7.6 | 130.3 | 158.5 | 150 | 13.52 |
| ta | test | 2.2 | 38.9 | 28.0 | 20 | 4.04 |
|  | train | 8.0 | 135.1 | 101.5 | 145 | 14.41 |
|  | valid | 2.1 | 35.4 | 27.3 | 42 | 3.56 |

Table 15: Statistics of Indic track dataset. #Lines and #Tokens (.en & .lang) are in terms of thousands($K$). All the data in the above table is approximated.

**English-Hindi**: Hindi is the third most spoken language in the world, with 615 million speakers. It belongs to the Indo-Aryan language family, mainly spoken in India. It is also the official language of India, written in devanagiri script. The data contains English speech, English text (transcripts), and Hindi text (translations). The speech in English language is 103.5 hours and the text in Hindi language is 37K lines.

**English-Bengali**: Bengali is the 7th most spoken language in the world, with 228 million speakers. It belongs to the Indo-Aryan language family, spoken in the Bengal region of South Asia. It is also the official language of Bangladesh, written in Bengali-Assamese script. The data contains English speech, English texts (transcripts), and Bengali texts (translations). The speech in English language is 13.59 hours and the text in Bengali language is 6.9K lines.

**English-Tamil**: Tamil is one of the classical languages of India, spoken by 90.8 million speakers. It belongs to dravidian language family, spoken by the tamil people of South Asia. It is the official language of Tamil Nadu state of India, written in Brahmi script. The data contains English speech, English texts (transcripts), and Tamil texts (translations). The speech in English language is 22.01 hours and the text in Tamil language is 8K lines.

**Metrics**: Case-sensitive detokenized BLEU using sacreBLEU (Post, 2018) is used to report the performance of all the submissions.

### 8.3 Submissions

There were four teams participating in this inaugural task: Research team from National Institute of Information and Communications Technology of Japan (NICT), the Voice Intelligence Team of Samsung (SRI-B), the Huawei Translation Service Center (HW-TSC), and a team from National Institute of Technology Kurukshetra, India (NITKKR). The participants submitted their result under various constraints, including end-to-end constrained, unconstrained, cascaded end-to-end, and unconstrained approaches. Below, we provide an overview of each team's approach and their results.

**NICT**: Their submission included cascaded and end-to-end approach in unconstrained setting for all the language pairs. The cascaded system involves fine-tuning the Whisper model for ASR and fine-tuning the IndicTrans2 model for MT. This dual fine-tuning aimed to address the format mismatch between spoken and written language. For the end-to-end syatem, the IndicTrans2 model is used to generate pseudo translation data, which replaced the gold transcription data for fine-tuning the Whisper model. This strategy aimed to distill knowledge from a stronger translation model and ensure consistent formatting. In stage 1, Whisper is fine-tuned using English transcription and Indic language translation. Stage 2 involves generation of pseudo translations for all English transcriptions, and fine-tuning Whisper using English audio and the pseudo translations. During inference, the fine-tuned Whisper model performed direct end-to-end speech translation.

**HW-TSC**: The submission included implementation of cascaded approach in the unconstrained setting. It involves Whisper-large-v3 model for Automatic Speech Recognition (ASR) and a Transformer model for Machine Translation (MT). For MT, strategies like LaBSE for parallel corpus filtering, data diversification using multiple model predictions, forward and back translation for data augmentation, domain fine-tuning with scored data selection, and regularized dropout for enhanced training efficiency are used. The base architecture is from FAIRSEQ toolkit (Wang et al., 2020b) with hyperparameters of 2048 as batch size, learning rate of 5e-4, label-smoothing-cross-entropy loss with label smoothing of 0.1, 4000 warmup steps, and Adam optimizer settings ($\beta_1 = 0.9$, $\beta_2 = 0.98$). During inference, a beam size of 4 and length penalties of 1.0 is applied to optimize translation outputs.

**SRI-B**: The submission included end-to-end approach in both constrained and unconstrained setting. In the constrained setting the base architecture used is from FAIRSEQ toolkit (Wang et al., 2020b). Pre-processing involves the extraction of 80 channel log mel-filter bank features with a window size of 25ms and SpecAugment for data augmentation. The s2t_conformer among fairseq's built-in architectures for speech-to-text translation is used. It consist of 16 encoder layers and 6 decoder layers with label-smoothed cross-entropy loss and the Adam optimizer with a learning rate of 2e-3 to train the models. Under the unconstrained setting, the method involves using the pre-trained SeamlessM4T v2 from Meta, a multi-lingual end-to-end model designed for various languages. The pre-trained multi-lingual model is used to directly generate text in Indic languages directly from English for evaluation.

**NITKKR**: The submission adopts cascaded approach in unconstrained setting to solve the task. It begins with audio preprocessing and transcription, utilizing ResembleAI for noise reduction, distortion restoration, and speech bandwidth enhancement. The processed audio is then fed into OpenAI's Whisper model for real-time ASR. Subsequently, MT models are applied: Helsinki-NLP's OPUS-MT for translating English to Hindi, and Facebook's Multilingual BART (MBART) for both English to Tamil and English to Bengali translations.

### 8.4 Results

Scores on the test set of all submissions are calculated using automatic metrics and the respective settings are presented in Table 16. In the following section, we discuss results from each direction of languages.

#### 8.4.1 En-Hi

**Unconstrained Setting**: In the E2E approach, NICT achieved a BLEU score of 33.02, significantly outperforming SRI-B, which scored 21.63. This superior performance by NICT can be attributed to their robust use of pseudo translation data aimed to distill knowledge from a stronger translation model to ensure consistent formatting. In the cascaded approach, NICT again led with a remarkable 60.54 BLEU score, significantly higher than HW-TSC at 47.14 and NITKKR at 19.77. The cascaded approach by NICT utilized the strengths of pretraining the ASR and MT model to address the format mismatch problem which leads to maximizing the performance.

| Language | Setting | Approach | Team ID | BLEU |
|---|---|---|---|---|
| En-Hi | Unconstrained | E2E | NICT | 33.02 |
| | | | SRI-B | 21.63 |
| | | Cascaded | NICT | **60.54** |
| | | | HW-TSC | 47.14 |
| | | | NITKKR | 19.77 |
| | Constrained | E2E | SRI-B | 29.76 |
| En-Bn | Unconstrained | E2E | NICT | 10.79 |
| | | | SRI-B | 18.13 |
| | | Cascaded | NICT | **52.63** |
| | | | HW-TSC | 35.04 |
| | | | NITKKR | 4.46 |
| | Constrained | E2E | SRI-B | 2 |
| En-Ta | Unconstrained | E2E | NICT | 13.46 |
| | | | SRI-B | 11.93 |
| | | Cascaded | NICT | **39.84** |
| | | | HW-TSC | 30.79 |
| | | | NITKKR | 11.76 |
| | Constrained | E2E | SRI-B | 0.81 |

Table 16: Results on all language pairs and setting from all the submissions.

**Constrained**: In the E2E approach, there was one submission by SRI-B, which achieved a BLEU score of 29.76.

### 8.4.2 En-Bn

**Unconstrained**: SRI-B with a BLEU score of 18.13 beats NICT which scored 10.79 when implementing the E2E approach. In the cascaded approach, NICT scored the highest with 52.63 BLEU, compared to HW-TSC at 35.04 and NITKKR at 4.46. The same strategy from En-Hi allowed NICT to excel in this category, demonstrating the effectiveness of their cascaded approach.

**Constrained**: For the E2E approach, SRI-B scored a BLEU of 2 demonstrating the challenges of the constrained setting in this language pair.

### 8.4.3 En-Ta

**Unconstrained**: NICT led with a BLEU score of 13.46, while SRI-B scored 11.93 for the models using E2E approach. NICT's consistent use of Whisper for ASR and their robust translation models contributed to their leading position. For teams using the cascaded approach, NICT again achieved the highest BLEU score of 39.84, followed by HW-TSC at 30.79 and NITKKR at 11.76. The result could be explained due to the method of addressing the format mismatch problem by NICT already mentioned above.

**Constrained**: In this setting there is one submission using the E2E approach, by SRI-B. They achieve a score of 0.81, which shows the limitations on this setting and language pair. The low score could be explained due to limited data and the morphologically complex structure of the Tamil language.

## 8.5 Conclusion

This is the first time that a speech-to-translation task is presented for the Indic track as one of the IWSLT tasks. The results presented in the work establish an important benchmark for the end-to-end as well as cascade models for both the constrained and unconstrained conditions. This work highlights a major performance gap between the end-to-end and the cascade models. Also, a noteworthy gap is seen in the performance with the unconstrained data and pretrained models are used. We plan to include more data and more Indic languages in the next edition.

## Acknowledgements

The FBK team (Roldano Cattoni, Mauro Cettolo, Matteo Negri and Sara Papi) is co-funded by the European Union under the project *AI4Culture: An AI platform for the cultural heritage data space* (Action number 101100683).

Atul Kr. Ojha and John P. McCrae would like to thank Science Foundation Ireland (SFI) under Grant Number SFI/12/RC/2289_ P2 Insight_2 and thanks to RTÉ/TG4 for sharing the Irish speech data. We would also like to thank Panlingua Language Processing LLP for providing the Marathi-Hindi, and Bhojpuri-Hindi speech translation data and for their support.

The work by Tsz Kin Lam and Barry Haddow was funded by UK Research and Innovation (UKRI) under the UK government's Horizon Europe funding guarantee (grant number 10039436: UTTER).

The work by Mateusz Krubiński, Petr Zemánek, Adam Pospíšil, and Pavel Pecina was funded by the European Commission via its H2020 Program (contract no. 870930: WELCOME).

The work by University of Malta was supported through H2020 EU Funded LT-Bridge Project (GA 952194) and DFKI for access to the Virtual Laboratory.

Brian Thompson's contributions to this work were conducted outside of, and are unrelated to, his employment at Amazon.

Ondřej Bojar would like to acknowledge the support from the 19-26934X (NEUREM3) grant of the Czech Science Foundation. The work of Peter Polák has been supported by the Ministry of Education, Youth and Sports of the Czech Republic, Project No. LM2023062 (LINDAT/CLARIAH-CZ). Dávid Javorský has been supported by the grant 272323 of the Grant Agency of Charles University.

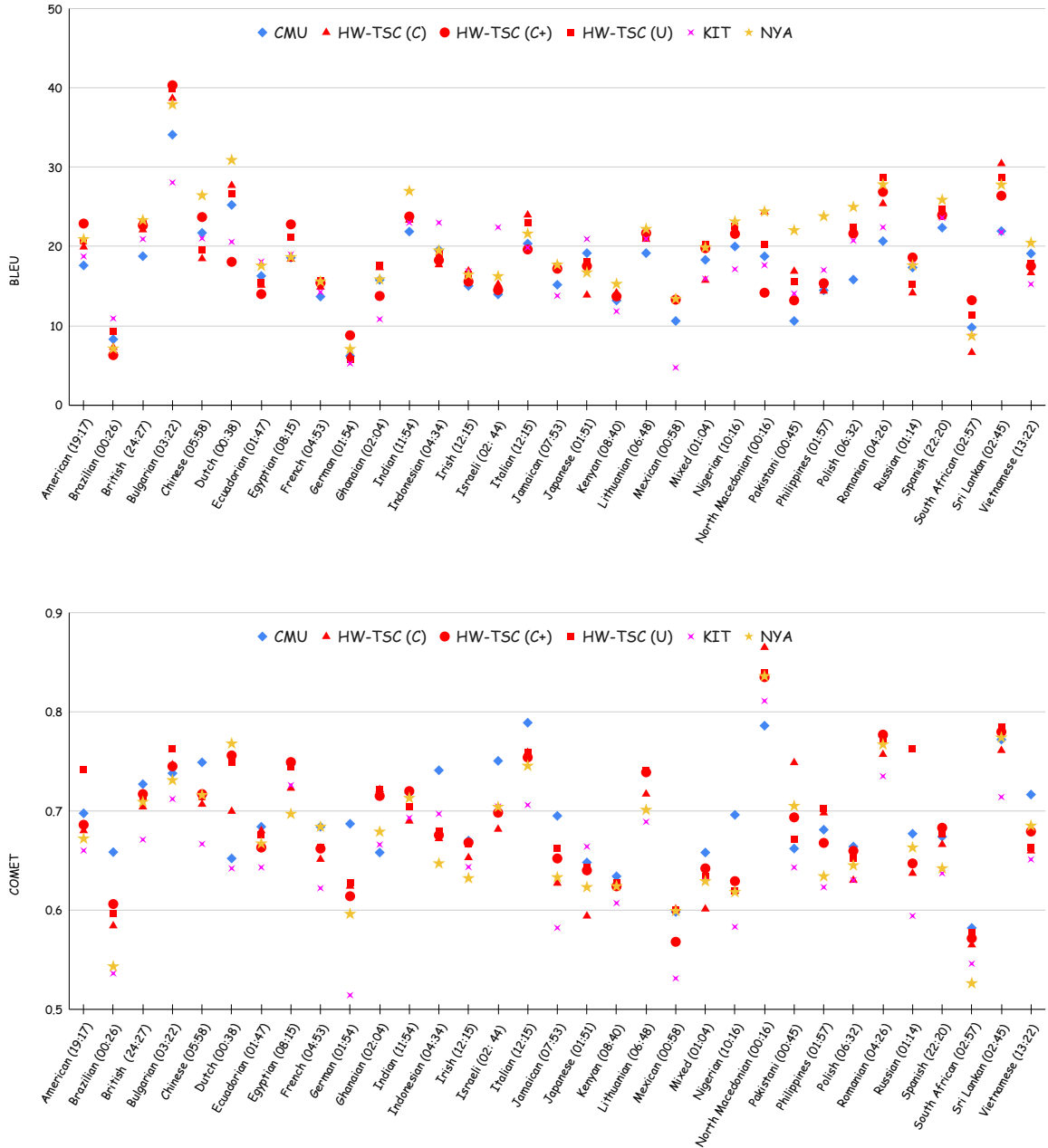

Figure 1: Performance in BLEU (up) and COMET (down) across a wide range of accents. The audio duration for each accent is denoted in a "(minutes:seconds)" format. The macro-average across accents are 18.7 BLEU and 0.679 COMET.

# Appendix A. Human Evaluation

## A Human Evaluation

Human evaluation included MQM for the English-to-Japanese simultaneous speech translation task (A.1), as well as direct assessment for offline, simultaneous, and subtitling tasks (A.2).

### A.1 MQM-based Human Evaluation for the English-to-Japanese Simultaneous Task

For the English-to-Japanese Simultaneous Translation Task, we conducted a human evaluation using a variant of Multidimensional Quality Metrics (MQM; Lommel et al., 2014). MQM has been used in recent MT evaluation studies (Freitag et al., 2021a) and WMT Metrics shared task (Freitag et al., 2021b). For the evaluation of Japanese translations, we used *JTF Translation Quality Evaluation Guidelines* (JTF, 2018), distributed by Japan Translation Federation (JTF). The guidelines are based on MQM but include some modifications in consideration of the property of the Japanese language.

We hired a Japanese-native professional interpreter as the evaluator. The evaluator checked translation hypotheses along with their source speech transcripts and chose the corresponding error category and severity for each translation hypothesis on a spreadsheet. Here, we asked the evaluator to focus only on *Accuracy* and *Fluency* errors, because other types of errors in Terminology, Style, and Locale convention would not be so serious in the evaluation of simultaneous translation. Finally, we calculated the cumulative error score for each system based on the error weighting presented by Freitag et al. (2021a), where *Critical* and *Major* errors have the same level of error scores. The results are shown in Table 17.

### A.2 Direct Assessment

For the offline translation track (Section 2), simultaneous translation track (Section 3), and subtitling track (Section 4), we conducted a human evaluation of primary submissions based on a random selection of 1000 segments from each test set. Human graders were asked for direct assessment (DA) (Graham et al., 2013; Cettolo et al., 2017; Akhbardeh et al., 2021), expressed through scores between 0 and 100.

#### A.2.1 Automatic Segmentation

In the case of offline and subtitling tracks, we collected segment-level annotations based on the re-segmentated test data (see Section 2). Because we did not want issues from the segmentation to influence scores negatively, we followed Sperber et al. (2024) and provided translators not only with the source sentence and system translation, but also with the system translation of the previous and following segments. Annotators were then instructed as follows: *"Sentence boundary errors are expected and should not be factored in when judging translation quality. This is when the translation appears to be missing or adding extra words but the source was segmented at a different place. To this end, we have included the translations for the previous and next sentences also. If the source and translation are only different because of sentence boundary issues, do not let this affect your scoring judgement. Example for a clear case for a good translation suffering only from sentence boundary issues that should not result in a poor score:*
*Source: \*you'll see that there's actually\* a sign near the road.*
*Translation: ein Schild neben der Straße gibt.*

| Team | BLEU (on three talks) | Error score | # Errors | | |
|---|---|---|---|---|---|
| | | | Critical | Major | Minor |
| NAIST | 17.2 | 27.4 | 0 | 3 | 16 |
| HW-TSC | 20.6 | 50.2 | 0 | 8 | 12 |
| FBK | 11.4 | 130.5 | 1 | 21 | 25 |

Table 17: Human evaluation results on two talks (107 lines) in the English-to-Japanese Simultaneous speech-to-text translation task. Error weights are 5 for Critical and Major errors and 1 for Minor errors.

*Previous sentence: Ich bin mir sicher, dass Sie nicht wissen, dass, wenn Sie weiter weitergehen, *Sie sehen – (Gelächter) – dass es tatsächlich**
*Next sentence: …."*

No video or audio context was provided. Segments were shuffled and randomly assigned to annotators to avoid bias related to the presentation order. Annotation was conducted by professional translators fluent in the source language and native in the target language.

### A.2.2 Subtitling Constraints

The subtitling task (Section 4) includes cases where systems compress translations in order to match subtitling constraints, e.g. filtering out non-relevant information present in the source. This is desired in subtitling and should therefore not be penalized in human evaluation. To this end, we provided annotators with the following instructions: *"When judging the translations, please consider that these are subtitles which are compressed translations of the original speech, not the translations of the subtitles in the source language. Thus, there may be significant differences in how the source and the target sentences are formulated. Subtitles are created independently for each language with the goal of good readability during the short time period when they are displayed on screen. Readability in terms of number of characters per second may differ between the source (English) and target (German). Please take this into account. The translation should convey the same meaning as the source sentence but may omit information that is not very important for getting the main message of the sentence across. It is OK if the sentence is shortened this way in order to fulfil the readability constraints."*

### A.2.3 Computing rankings

System rankings are produced from the average DA scores computed from the average human assessment scores according to each individual annotator's mean and standard deviation, similarly to Akhbardeh et al. (2021). Ranks are established according to Wilcoxon rank-sum statistical significance test with $p < 0.05$. The below tables show the DA scores and rankings. Note that the guidelines are different for offline, simultaneous, and subtitling tasks. This makes results not directly comparable across tasks, and we consequently only present within-task rankings here. Within each of the tasks (only the offline and subtitling English-to-German have more domains), all the outputs were assessed in one annotation run, distributing the scoring items randomly to annotators across domains, with all annotators most likely seeing all the domains. This allows us to treat the DA scores across domains in a given task as comparable, so we present them in the same table.

Table 18: Offline task, **English to German**

| System | All Rank | All DA | TED Rank | TED DA | ITV Rank | ITV DA | Accent Rank | Accent DA | Peloton Rank | Peloton DA |
|---|---|---|---|---|---|---|---|---|---|---|
| HWTSC-LLM | 1 | 84.8 | 1-2 | 94.9 | 1-2 | 84.7 | 1-4 | 76.1 | 1-4 | 82.6 |
| HWTSC | 2-3 | 84.2 | 3-5 | 92.8 | 1-3 | 84.0 | 1-4 | 76.8 | 1-4 | 81.6 |
| CMU | 2-4 | 83.3 | 3-5 | 92.5 | 2-3 | 83.1 | 1-4 | 75.4 | 1-4 | 81.2 |
| NYA | 3-4 | 81.0 | 1-2 | 94.7 | 4 | 73.9 | 1-4 | 77.9 | 1-4 | 80.2 |
| KIT | 5 | 76.7 | 3-5 | 91.8 | 5 | 69.3 | 5 | 72.8 | 5 | 74.6 |

Table 19: Offline task, **English to Japanese**

| System | TED Rank | DA |
|---|---|---|
| HWTSC | 1-3 | 75.4 |
| HWTSC-LLM | 1-2 | 74.7 |
| NYA | 2-4 | 72.8 |
| CMU | 3-4 | 72.9 |

Table 20: Offline task, **English to Chinese**

| System | TED Rank | DA |
|---|---|---|
| HWTSC-LLM | 1 | 78.9 |
| NYA | 2-3 | 77.2 |
| HWTSC | 2-4 | 76.5 |
| CMU | 3-4 | 75.8 |

Table 21: Simultaneous task, English to German

| System | TED Rank | DA |
|---|---|---|
| CMU | 1 | 87.3 |
| HWTSC | 2 | 86.0 |
| FBK | 3-4 | 84.2 |
| NAIST | 3-4 | 83.4 |

Table 22: Simultaneous task, English to Japanese

| System | TED Rank | DA |
|---|---|---|
| NAIST | 1 | 77.4 |
| HWTSC | 2 | 75.4 |
| FBK | 3 | 71.7 |

Table 23: Simultaneous task, English to Chinese

| System | TED Rank | DA |
|---|---|---|
| HWTSC | 1-2 | 80.0 |
| NAIST | 1-2 | 79.2 |
| FBK | 3 | 76.1 |

Table 24: Subtitling task, English to German. *All* combines the ITV and Peloton DA scores

|             | All  |      | ITV  |      | Peloton |      |
| System      | Rank | DA   | Rank | DA   | Rank    | DA   |
|-------------|------|------|------|------|---------|------|
| HWTSC       | 1    | 72.2 | 1    | 73.0 | 1-2     | 71.3 |
| AppTek      | 2-3  | 68.2 | 2    | 69.3 | 3       | 67.3 |
| FBK-cascade | 2-3  | 66.3 | 3    | 62.2 | 1-2     | 71.5 |
| FBK-direct  | 4    | 52.8 | 4    | 46.5 | 4       | 61.2 |

# Appendix B. Automatic Evaluation Results and Details

## B.1 Offline SLT

- Systems are ordered according to the COMET score (denoted by COMET, the third column).
- The "Joint" column is computed by averaging the scores of the 4 test sets, aka macro-averaging.
- The "D" column indicates the data condition in which each submitted run was trained, namely: Constrained (C), Constrained$^{+LLM}$ (C$^+$), Unconstrained (U).
- All systems are based on cascade architecture.

| System | D | Joint | | TED 2024 | | ITV | | Peloton | | Accent | |
| --- | --- | --- | --- | --- | --- | --- | --- | --- | --- | --- | --- |
| | | COMET | BLEU | COMET | BLEU | COMET | BLEU | COMET | BLEU | COMET | BLEU |
| CMU | U | 0.743 | 18.3 | 0.862 | 25.7 | 0.735 | 17.3 | 0.670 | 11.5 | 0.705 | 18.5 |
| HW-TSC | C$^+$ | 0.731 | 19.3 | 0.851 | 27.4 | 0.728 | 17.2 | 0.652 | 11.9 | 0.691 | 20.7 |
| HW-TSC | U | 0.727 | 19.1 | 0.849 | 27.1 | 0.723 | 17.3 | 0.646 | 11.0 | 0.690 | 20.8 |
| HW-TSC | C | 0.717 | 18.5 | 0.841 | 26.6 | 0.712 | 16.7 | 0.637 | 10.4 | 0.678 | 20.2 |
| NYA | U | 0.695 | 19.5 | 0.837 | 28.1 | 0.648 | 15.8 | 0.616 | 12.2 | 0.677 | 21.7 |
| KIT | C$^+$ | 0.677 | 17.5 | 0.832 | 27.5 | 0.618 | 13.2 | 0.600 | 10.2 | 0.656 | 19.1 |

Table 25: Official results of the automatic evaluation for the Offline Speech Translation Task, **English to German**.

| System | D | TED 2023 | | EMPAC | | ACL | |
| --- | --- | --- | --- | --- | --- | --- | --- |
| | | COMET | BLEU | COMET | BLEU | COMET | BLEU |
| CMU | U | 0.858 | 27.2 | 0.820 | 16.2 | 0.837 | 31.5 |
| HW-TSC | U | 0.849 | 32.6 | 0.799 | 17.4 | 0.823 | 38.3 |
| HW-TSC | C$^+$ | 0.844 | 29.0 | 0.802 | 18.4 | 0.825 | 38.2 |
| HW-TSC | C | 0.843 | 32.8 | 0.792 | 17.1 | 0.808 | 37.0 |
| NYA | U | 0.837 | 29.8 | 0.756 | 17.2 | 0.826 | 45.5 |
| KIT | C$^+$ | 0.831 | 28.7 | 0.723 | 15.2 | 0.781 | 35.1 |
| Best 2023 | | 0.821 | 30.2 | 0.382 | 16.9 | 0.801 | 41.1 |

Table 26: Official results of the automatic evaluation for the Offline Speech Translation Task on progress test sets, **English to German**.

| System | D | TED 2024 | | TED 2023 | | ACL | |
| --- | --- | --- | --- | --- | --- | --- | --- |
| | | COMET | BLEU | COMET | BLEU | COMET | BLEU |
| HW-TSC | U | 0.853 | 23.6 | 0.856 | 23.1 | 0.868 | 31.8 |
| HW-TSC | C$^+$ | 0.851 | 23.1 | 0.856 | 22.2 | 0.839 | 32.5 |
| CMU | U | 0.841 | 18.3 | 0.850 | 17.9 | 0.849 | 19.1 |
| HW-TSC | C | 0.839 | 23.9 | 0.831 | 24.3 | 0.839 | 28.0 |
| NYA | U | 0.812 | 20.1 | 0.822 | 21.0 | 0.861 | 39.9 |

Table 27: Official results of the automatic evaluation for the Offline Speech Translation Task on official test set and progress test sets, **English to Japanese**.

| System  | D    | TED 2024 |      | TED 2023 |      | ACL   |      |
|---------|------|----------|------|----------|------|-------|------|
|         |      | COMET    | BLEU | COMET    | BLEU | COMET | BLEU |
| HW-TSC  | U    | 0.845    | 37.0 | 0.834    | 36.3 | 0.857 | 50.8 |
| HW-TSC  | C$^+$ | 0.842   | 36.2 | 0.831    | 35.8 | 0.855 | 49.8 |
| CMU     | U    | 0.834    | 31.5 | 0.827    | 30.6 | 0.853 | 43.1 |
| HW-TSC  | C    | 0.824    | 38.3 | 0.810    | 37.3 | 0.833 | 52.4 |
| NYA     | U    | 0.823    | 40.4 | 0.814    | 39.1 | 0.855 | 59.1 |

Table 28: Official results of the automatic evaluation for the Offline Speech Translation Task on official test set and progress test sets, **English to Chinese**.

# Translation Guidelines

In this task, we aim to obtain high quality German translations of the English transcripts. The transcripts (inside the "transcripts.txt" file) contain conversations between friends talking about a daily topic, e.g. hobbies and vacation. There are **76** conversations (recordings) in total. In each conversation, there are only two speakers, but the same pair of speakers may appear in another set(s) of conversations, see the list below. **The content of each recording is independent of each other, so they could be translated independently**. For each source sentence (line) to be translated, we provide metadata, such as the recording id, speaker id, the audio file and the utterance number. The utterance number indicates its order in the conversation. It begins from 0 (which is not included in the transcripts required for translation) and stands for the beginning of the conversation. In general, most recordings start from an utterance number of 15.

The general translation guidelines are:
- All translations should be **"from scratch", without post-editing from Machine Translation.** We can detect post editing so will reject translations that are post-edited.
- Translators should **preserve the line structure of the source file**. By this we mean that they should not add or remove line-breaks, and each line is English should correspond to a line of German. Note that each line of the source file corresponds to one audio file.
- We need the **translations to be returned in the same format**. If you prefer to receive the text in a different format, then please let us know as we may be able to accommodate it.
- Translators should **avoid inserting parenthetical explanations** into the translated text and obviously **avoid losing any pieces of information from the source text**. We will check a sample of the translations for quality, and we will check the entire set for evidence of post-editing.

Since it is a conversation between friends, please pay attention to the below:
- You might need to use the context before and/or after the utterance to translate.
- **[Important]** There are disfluencies in the transcripts, including but not limited to, hesitation, repetitions, and correction. **We expect to have fluent and faithful translations. These disfluencies in the transcripts might be helpful for your translation, but they are not required as long as the meaning is clear. Please avoid word-by-word translation of them.**
    a. In general, please focus on the core meaning in the translation. You might rephrase or remove the redundant parts in the transcripts if necessary, e.g., repetitions.
    b. For Hesitation, some examples are below, please do **NOT** include them in the translation. We keep them on the transcripts as it might help signal a "pause" in the utterance.

Examples of disfluencies:
- Hesitation:
    a. List of possible tokens: {"ACH", "AH", "EEE", "EH", "ER", "EW", "HA", "HEE","HM", "HUH", "MM", "OOF", "UH", "UM", "HMM"}
    b. Example: "YEAH I KNOW ~~UM~~ WAIT WHAT WAS I GONNA SAY ~~UM~~ SO DO YOU WANNA ASK THE QUESTION NOW"?
- Repetitions:

a. "WELL ACTUALLY ARE THEY LIKE ALL THESE ~~ALL THESE ALL THESE~~ DUMPLINGS OF EASTERN EUROPEAN ORIGIN"

**A note on the recording_id**

There are 76 conversations / recordings in total, but the same pair of speakers may show up in another conversation(s) (122 speakers in total). In spite of the same pair of speakers, the contents in each of these conversations are also independent of each other. These conversations have their id extended by "_PX" where "X" is a number. Below is the list of recodings that have "_PX" in their names:

- EDACC-C23_P1, EDACC-C23_P2
- EDACC-C32_P1, EDACC-C32_P2
- EDACC-C33_P1
- EDACC-C40_P1, EDACC-C40_P2, EDACC-C40_P3
- EDACC-C43_P1
- EDACC-C46_P1, EDACC-C46_P2
- EDACC-C05_P0, EDACC-C05_P1
- EDACC-C29_P1, EDACC-C29_P2
- EDACC-C31_P1, EDACC-C31_P2
- EDACC-C38_P1, EDACC-C38_P2
- EDACC-C35_P1, EDACC-C35_P2, EDACC-C35_P3
- EDACC-C36_P1, EDACC-C36_P2
- EDACC-C37_P1, EDACC-C37_P2
- EDACC-C47_P1, EDACC-C47_P2
- EDACC-C57_P1, EDACC-C57_P2

## B.2 Simultaneous SLT

| Team | BLEU | LAAL | AL | AP | DAL | ATD |
|---|---|---|---|---|---|---|
| HW-TSC | 26.39 | 2.17 (4.19) | 1.92 (4.07) | 0.919 (1.66) | 3.10 (7.37) | 2.18 (5.31) |
| CMU | 24.65 | 2.21 (3.57) | 2.01(3.45) | 0.87 (1.24) | 3.04 (4.73) | 2.22 (3.22) |
| NAIST | 23.37 | 2.30 (3.33) | 2.05 (3.17) | 0.91 (1.22) | 3.03 (4.53) | 2.23 (3.12) |
| FBK | 21.18 | 2.00 (3.03) | 1.71 (2.84) | 0.92 (1.24) | 2.52 (3.77) | 2.02 (2.49) |

Table 29: Simultaneous Speech-to-Text Translation, English to German. Except for AP, the latency is measured in seconds. Numbers in brackets are computation aware latency.

| Team | BLEU | LAAL | AL | AP | DAL | ATD |
|---|---|---|---|---|---|---|
| HW-TSC | 34.23 | 2.10 (3.93) | 2.00 (3.89) | 0.78 (1.42) | 3.05 (7.45) | 0.94 (4.24) |
| NAIST | 29.33 | 2.36 (3.19) | 2.24 (3.11) | 0.79 (1.06) | 3.01 (4.51) | 1.04 (1.81) |
| FBK | 25.20 | 2.73 (4.43) | 2.61 (4.16) | 0.84 (1.17) | 3.61 (5.44) | 1.09 (2.42) |

Table 30: Simultaneous Speech-to-Text Translation, English to Chinese. Except for AP, the latency is measured in seconds. Numbers in brackets are computation aware latency.

| Team | BLEU | LAAL | AL | AP | DAL | ATD |
|---|---|---|---|---|---|---|
| HW-TSC | 19.394 | 2.44 (4.10) | 2.39 (4.01) | 0.77 (1.28) | 3.35 (7.03) | 0.74 (3.44) |
| NAIST | 17.954 | 2.39 (3.41) | 2.31 (3.37) | 0.79 (1.14) | 3.08 (5.21) | 0.56 (1.68) |
| FBK | 12.136 | 2.15 (3.74) | 2.07 (3.70) | 0.72 (1.18) | 2.85 (5.53) | 0.59 (2.25) |

Table 31: Simultaneous Speech-to-Text Translation, English to Japanese. Except for AP, the latency is measured in seconds. Numbers in brackets are computation aware latency.

| Team    | BLEU  | LAAL        | AL          | AP          | DAL          | ATD         |
|---------|-------|-------------|-------------|-------------|--------------|-------------|
| BENCH-2 | 29.93 | 2.28        | 1.95        | 0.78        | 3.03         | 2.75        |
| BENCH-1 | 29.43 | 2.35        | 2.02        | 0.82        | 3.13         | 2.78        |
| FBK     | 29.20 | 2.55 (3.92) | 2.14 (3.65) | 0.93 (1.24) | 3.20 (4.67)  | 2.75 (3.29) |
| HW-TSC  | 27.11 | 2.00 (5.11) | 1.53 (4.86) | 0.89 (2.28) | 3.27 (11.03) | 2.63 (8.38) |
| BENCH-0 | 26.85 | 3.34        | 3.09        | 0.75        | 3.99         | 3.39        |

Table 32: Simultaneous Speech-to-Text Translation, Czech to English. Except for AP, the latency is measured in seconds. Numbers in brackets are computationally-aware latency. BENCH-$N$ represents ORGANIZER'S BENCHMARK, with $N$ indicating the number of previously translated segments used as a Whisper prompt to provide the model with the context.

| Target Language     | Team   | ASR BLEU | Start Offset | End Offset | ATD  |
|---------------------|--------|----------|--------------|------------|------|
| English to German   | HW-TSC | 23.33    | 2.00         | 4.30       | 3.22 |
| English to Japanese | HW-TSC | 17.37    | 2.36         | 3.41       | 3.31 |
|                     | NAIST  | 14.35    | 2.39         | 4.20       | 4.18 |
| English to Chinese  | HW-TSC | 28.97    | 2.04         | 2.99       | 3.11 |
| Czech to English    | HW-TSC | 25.93    | 1.58         | 3.52       | 3.67 |

Table 33: Simultaneous Speech-to-Speech from English Speech. The latency is measured in seconds. The BLEU scores are computed based on transcript from the default Whisper (Radford et al., 2023) ASR model (large) for each language direction.

## B.3 Automatic Subtitling

| Team | Condition | System | Domain | Subtitle quality SubER | Translation quality Bleu | ChrF | Bleurt | Subtitle compliance CPS | CPL | LPB |
|---|---|---|---|---|---|---|---|---|---|---|
| APPTEK | U | cntrstv2 | ALL | 70.34 | 17.45 | 41.77 | .4746 | 73.25 | 100.00 | 98.78 |
| | | | ted | 60.55 | 24.70 | 53.00 | .5823 | 86.89 | 100.00 | 97.27 |
| | | | itv | 72.19 | 16.47 | 39.12 | .4575 | 65.46 | 100.00 | 99.18 |
| | | | pltn | 77.68 | 10.22 | 32.38 | .3910 | 83.70 | 100.00 | 99.14 |
| APPTEK | U | prmry | ALL | 71.01 | 17.54 | 42.82 | .4842 | 73.94 | 100.00 | 99.78 |
| | | | ted | 63.03 | 23.35 | 54.03 | .5904 | 79.67 | 100.00 | 99.33 |
| | | | itv | 72.38 | 16.98 | 40.42 | .4683 | 69.23 | 100.00 | 99.92 |
| | | | pltn | 77.45 | 10.17 | 32.46 | .3981 | 83.03 | 100.00 | 99.80 |
| APPTEK | U | cntrstv1 | ALL | 71.52 | 17.48 | 43.28 | .4874 | 67.18 | 100.00 | 96.73 |
| | | | ted | 63.97 | 23.13 | 55.09 | .6024 | 73.91 | 100.00 | 91.81 |
| | | | itv | 72.79 | 16.88 | 40.62 | .4689 | 61.24 | 100.00 | 97.88 |
| | | | pltn | 77.64 | 10.26 | 32.70 | .3987 | 79.17 | 100.00 | 98.40 |
| FBK-AI4C$_{DIR}$ | C | prmry | ALL | 73.99 | 13.48 | 36.12 | .3775 | 76.19 | 88.86 | 99.99 |
| | | | ted | 57.50 | 25.79 | 54.78 | .6114 | 83.10 | 83.69 | 100.00 |
| | | | itv | 78.90 | 9.67 | 28.43 | .2911 | 70.45 | 90.04 | 99.97 |
| | | | pltn | 80.68 | 7.71 | 30.45 | .3542 | 82.16 | 92.77 | 100.00 |
| HW-TSC | U | cntrstv2 | ALL | 74.44 | 16.70 | 41.78 | .5008 | 86.40 | 60.18 | 100.00 |
| | | | ted | 69.44 | 22.40 | 50.60 | .5513 | 93.98 | 37.83 | 100.00 |
| | | | itv | 74.72 | 16.08 | 40.18 | .5031 | 82.84 | 65.55 | 100.00 |
| | | | pltn | 80.26 | 11.11 | 32.89 | .4284 | 90.62 | 66.12 | 100.00 |
| FBK-AI4C$_{CSC}$ | U | prmry | ALL | 75.56 | 16.23 | 40.10 | .4503 | 64.64 | 91.79 | 100.00 |
| | | | ted | 63.26 | 22.94 | 53.70 | .5872 | 79.99 | 89.52 | 100.00 |
| | | | itv | 79.92 | 14.86 | 35.16 | .4048 | 54.20 | 91.12 | 100.00 |
| | | | pltn | 78.34 | 11.30 | 34.13 | .4202 | 76.52 | 96.99 | 100.00 |
| HW-TSC | U | prmry | ALL | 75.60 | 16.62 | 42.64 | .5066 | 67.92 | 57.34 | 100.00 |
| | | | ted | 70.27 | 22.09 | 50.97 | .5556 | 80.09 | 36.44 | 100.00 |
| | | | itv | 76.04 | 16.09 | 41.34 | .5098 | 61.72 | 61.80 | 100.00 |
| | | | pltn | 81.35 | 11.13 | 33.56 | .4332 | 76.40 | 64.93 | 100.00 |
| HW-TSC | U | cntrstv1 | ALL | 77.11 | 16.52 | 43.00 | .5148 | 28.67 | 62.64 | 100.00 |
| | | | ted | 70.48 | 22.06 | 51.00 | .5559 | 46.25 | 36.66 | 100.00 |
| | | | itv | 78.04 | 16.07 | 41.80 | .5194 | 19.80 | 66.38 | 100.00 |
| | | | pltn | 83.09 | 10.93 | 34.25 | .4467 | 40.61 | 74.57 | 100.00 |

Table 34: Subtitling Task: automatic evaluation scores on tst2024 en→de. *C* and *U* stand for *constrained* and *unconstrained* training condition, respectively; *prmry* and *cntrstv* for *primary* and *contrastive* systems. Ranking based on SubER scores on ALL domains.

| Team | Condition | System | Domain | Subtitle quality SubER | Translation quality Bleu | ChrF | Bleurt | Subtitle compliance CPS | CPL | LPB |
|---|---|---|---|---|---|---|---|---|---|---|
| APPTEK | U | prmry | ALL | 62.02 | 25.59 | 49.75 | .5268 | 82.42 | 100.00 | 99.94 |
| | | | ted | 45.73 | 39.29 | 63.86 | .6995 | 88.05 | 100.00 | 99.76 |
| | | | itv | 66.80 | 21.37 | 44.35 | .4761 | 79.18 | 100.00 | 99.98 |
| | | | pltn | 73.55 | 15.45 | 41.43 | .4728 | 86.83 | 100.00 | 100.00 |
| FBK-AI4C$_{\text{CSC}}$ | U | prmry | ALL | 63.01 | 26.60 | 49.64 | .5174 | 69.97 | 93.28 | 100.00 |
| | | | ted | 40.75 | 45.69 | 69.20 | .7500 | 83.42 | 90.31 | 100.00 |
| | | | itv | 70.82 | 18.92 | 40.17 | .4262 | 60.85 | 93.46 | 100.00 |
| | | | pltn | 74.17 | 16.18 | 44.42 | .5108 | 80.24 | 97.03 | 100.00 |
| APPTEK | U | cntrstv1 | ALL | 63.65 | 24.33 | 48.63 | .5152 | 75.98 | 100.00 | 98.52 |
| | | | ted | 47.71 | 37.61 | 62.68 | .6892 | 85.50 | 100.00 | 96.60 |
| | | | itv | 67.85 | 20.37 | 43.44 | .4668 | 70.82 | 100.00 | 98.98 |
| | | | pltn | 76.72 | 13.70 | 39.75 | .4533 | 82.37 | 100.00 | 99.14 |
| HW-TSC | U | cntrstv2 | ALL | 63.77 | 26.92 | 50.09 | .5453 | 91.43 | 62.67 | 100.00 |
| | | | ted | 49.64 | 42.35 | 64.55 | .6859 | 94.97 | 38.82 | 100.00 |
| | | | itv | 67.57 | 21.39 | 43.94 | .5045 | 90.09 | 69.19 | 100.00 |
| | | | pltn | 75.08 | 16.79 | 43.95 | .4999 | 92.26 | 66.20 | 100.00 |
| HW-TSC | U | prmry | ALL | 64.18 | 27.38 | 51.50 | .5554 | 74.80 | 60.42 | 100.00 |
| | | | ted | 48.93 | 44.20 | 66.12 | .6953 | 81.48 | 37.43 | 100.00 |
| | | | itv | 68.42 | 22.10 | 45.46 | .5159 | 71.28 | 66.28 | 100.00 |
| | | | pltn | 75.83 | 16.97 | 44.84 | .5071 | 79.90 | 65.38 | 100.00 |
| HW-TSC | U | cntrstv1 | ALL | 64.87 | 27.25 | 51.58 | .5583 | 33.42 | 66.14 | 100.00 |
| | | | ted | 49.02 | 44.18 | 66.11 | .6951 | 47.70 | 38.24 | 100.00 |
| | | | itv | 69.50 | 22.01 | 45.56 | .5183 | 25.92 | 71.22 | 100.00 |
| | | | pltn | 76.17 | 16.84 | 45.07 | .5150 | 44.09 | 75.58 | 100.00 |
| APPTEK | U | cntrstv2 | ALL | 66.25 | 22.25 | 47.74 | .4985 | 73.47 | 100.00 | 98.61 |
| | | | ted | 46.82 | 38.63 | 64.18 | .6853 | 84.53 | 100.00 | 96.48 |
| | | | itv | 72.12 | 17.40 | 41.15 | .4440 | 66.82 | 100.00 | 99.10 |
| | | | pltn | 79.46 | 12.60 | 39.25 | .4391 | 83.03 | 100.00 | 99.33 |
| FBK-AI4C$_{\text{DIR}}$ | C | prmry | ALL | 67.13 | 22.03 | 44.69 | .4277 | 76.00 | 90.35 | 100.00 |
| | | | ted | 39.86 | 45.63 | 69.63 | .7441 | 82.43 | 86.59 | 100.00 |
| | | | itv | 77.00 | 11.91 | 31.95 | .2986 | 70.61 | 92.60 | 100.00 |
| | | | pltn | 79.70 | 11.88 | 40.05 | .4329 | 82.26 | 89.58 | 100.00 |

Table 35: Subtitling Task: automatic evaluation scores on tst2024 en→es. *C* and *U* stand for *constrained* and *unconstrained* training condition, respectively; *prmry* and *cntrstv* for *primary* and *contrastive* systems. Ranking based on SubER scores on ALL domains.

| id | Team | System | de Bleurt | CPS | es Bleurt | CPS |
|---|---|---|---|---|---|---|
| 0 | subtitles to compress | | .1946 | 60.25 | .2136 | 69.97 |
| 1 | baseline | | .1720 | 100.00 | .1892 | 100.00 |
| 2 | FBK | primary | .1895 | 84.81 | .2063 | 90.66 |
| 3 | FBK | contrastive 1 | .1890 | 67.94 | .2113 | 75.74 |
| 4 | FBK | contrastive 2 | .1811 | 83.36 | .2033 | 87.48 |
| 5 | HW-TSC | primary | .1956 | 84.35 | .2101 | 91.42 |
| 6 | HW-TSC | contrastive 1 | .1967 | 79.97 | .2126 | 87.56 |
| 7 | HW-TSC | contrastive 2 | .2002 | 84.38 | .2102 | 91.44 |
| 8 | RACAI | primary | not submitted | | .1946 | 94.29 |

Table 36: Compression Task: automatic evaluation scores on German and Spanish subtitles.

| Team | Condition | System | Domain | Subtitle quality SubER | Translation quality Bleu | ChrF | Bleurt | Subtitle compliance CPS | CPL | LPB |
|---|---|---|---|---|---|---|---|---|---|---|
| APPTEK | U | cntrstv2 | ALL | 70.05 | 16.51 | 40.51 | .4730 | 70.46 | 100.00 | 98.87 |
|  |  |  | ted | 60.38 | 23.58 | 50.67 | .5808 | 82.29 | 100.00 | 97.50 |
|  |  |  | itv | 69.09 | 16.97 | 39.90 | .4718 | 65.00 | 100.00 | 99.03 |
|  |  |  | pltn | 78.02 | 9.96 | 34.41 | .4217 | 75.58 | 100.00 | 99.22 |
| APPTEK | U | prmry | ALL | 70.29 | 17.24 | 41.77 | .4813 | 72.13 | 100.00 | 99.84 |
|  |  |  | ted | 61.46 | 24.22 | 52.85 | .6012 | 77.30 | 100.00 | 99.45 |
|  |  |  | itv | 69.21 | 17.97 | 41.27 | .4790 | 67.64 | 100.00 | 99.96 |
|  |  |  | pltn | 77.99 | 10.46 | 34.67 | .4262 | 78.62 | 100.00 | 99.80 |
| APPTEK | U | cntrstv1 | ALL | 70.88 | 17.16 | 42.08 | .4846 | 65.08 | 100.00 | 97.13 |
|  |  |  | ted | 62.59 | 24.08 | 53.51 | .6097 | 70.12 | 100.00 | 92.51 |
|  |  |  | itv | 69.70 | 18.04 | 41.56 | .4818 | 59.96 | 100.00 | 97.91 |
|  |  |  | pltn | 78.45 | 10.29 | 34.76 | .4276 | 72.92 | 100.00 | 97.85 |
| HW-TSC | U | cntrstv2 | ALL | 72.37 | 17.69 | 41.75 | .5064 | 85.10 | 58.39 | 100.00 |
|  |  |  | ted | 62.79 | 26.33 | 52.40 | .5916 | 93.56 | 32.02 | 100.00 |
|  |  |  | itv | 71.35 | 18.10 | 41.39 | .5139 | 82.01 | 65.86 | 100.00 |
|  |  |  | pltn | 80.40 | 10.86 | 34.74 | .4508 | 88.04 | 54.55 | 100.00 |
| HW-TSC | U | prmry | ALL | 73.10 | 17.92 | 43.00 | .5156 | 65.44 | 55.51 | 100.00 |
|  |  |  | ted | 62.90 | 26.79 | 53.56 | .6013 | 78.54 | 30.30 | 100.00 |
|  |  |  | itv | 72.16 | 18.35 | 42.95 | .5244 | 60.15 | 62.37 | 100.00 |
|  |  |  | pltn | 81.38 | 10.91 | 35.46 | .4577 | 71.22 | 52.55 | 100.00 |
| FBK-AI4C$_{CSC}$ | U | prmry | ALL | 73.78 | 16.46 | 39.07 | .4454 | 61.44 | 93.04 | 100.00 |
|  |  |  | ted | 62.86 | 22.44 | 51.88 | .5910 | 76.28 | 90.67 | 100.00 |
|  |  |  | itv | 74.91 | 16.19 | 35.91 | .3996 | 54.70 | 92.97 | 100.00 |
|  |  |  | pltn | 78.38 | 10.59 | 36.09 | .4550 | 65.10 | 94.66 | 100.00 |
| FBK-AI4C$_{DIR}$ | C | prmry | ALL | 74.26 | 13.08 | 34.77 | .3742 | 72.75 | 89.35 | 99.96 |
|  |  |  | ted | 59.06 | 24.41 | 52.05 | .5996 | 79.52 | 83.97 | 99.94 |
|  |  |  | itv | 77.15 | 10.40 | 29.13 | .2939 | 68.73 | 91.00 | 99.97 |
|  |  |  | pltn | 78.03 | 9.41 | 33.39 | .4059 | 74.84 | 90.14 | 99.96 |
| HW-TSC | U | cntrstv1 | ALL | 74.34 | 17.80 | 43.57 | .5279 | 27.53 | 61.69 | 100.00 |
|  |  |  | ted | 63.21 | 26.61 | 54.29 | .6148 | 41.37 | 36.08 | 100.00 |
|  |  |  | itv | 74.12 | 18.23 | 43.42 | .5335 | 18.37 | 67.29 | 100.00 |
|  |  |  | pltn | 81.77 | 10.85 | 36.12 | .4751 | 41.44 | 61.99 | 100.00 |
| Submissions 2023 (here ALL={ted,itv,pltn}, while last year *eptv* was considered as well): | | | | | | | | | | |
| APPTEK | U | prmry | ALL | 70.23 | 15.10 | 37.39 | .4291 | 87.87 | 100.00 | 100.00 |
| MATESUB | U | prmry | ALL | 74.00 | 14.92 | 38.92 | .4579 | 84.47 | 99.26 | 100.00 |
| APPTEK | C | prmry | ALL | 77.14 | 12.40 | 33.17 | .3300 | 93.01 | 100.00 | 100.00 |
| FBK | C | prmry | ALL | 79.70 | 10.77 | 31.99 | .3016 | 69.23 | 83.72 | 99.99 |
| APPTEK | C | cntrstv | ALL | 83.75 | 9.33 | 29.28 | .2790 | 88.90 | 100.00 | 100.00 |

Table 37: Subtitling Task: automatic evaluation scores on tst2023 en→de. *C* and *U* stand for *constrained* and *unconstrained* training condition, respectively; *prmry* and *cntrstv* for *primary* and *contrastive* systems. Ranking based on SubER scores on ALL domains.

| Team | Condition | System | Domain | Subtitle quality SubER | Translation quality Bleu | ChrF | Bleurt | Subtitle compliance CPS | CPL | LPB |
|---|---|---|---|---|---|---|---|---|---|---|
| APPTEK | U | prmry | ALL | 63.97 | 23.25 | 47.46 | .5121 | 80.98 | 100.00 | 99.98 |
| | | | ted | 46.75 | 36.33 | 61.47 | .6889 | 88.92 | 100.00 | 99.84 |
| | | | itv | 66.39 | 22.17 | 45.42 | .4881 | 77.61 | 100.00 | 100.00 |
| | | | pltn | 71.61 | 15.47 | 40.75 | .4646 | 83.82 | 100.00 | 100.00 |
| HW-TSC | U | cntrstv2 | ALL | 64.72 | 25.00 | 49.02 | .5480 | 90.78 | 62.45 | 100.00 |
| | | | ted | 44.98 | 43.71 | 66.71 | .7240 | 94.76 | 33.30 | 100.00 |
| | | | itv | 67.35 | 22.17 | 45.13 | .5213 | 89.53 | 71.44 | 100.00 |
| | | | pltn | 73.73 | 17.20 | 43.05 | .5059 | 91.66 | 56.41 | 100.00 |
| APPTEK | U | cntrstv1 | ALL | 65.37 | 22.27 | 46.61 | .5007 | 74.41 | 100.00 | 98.91 |
| | | | ted | 48.98 | 34.49 | 60.17 | .6758 | 85.26 | 100.00 | 97.19 |
| | | | itv | 67.29 | 21.55 | 44.82 | .4784 | 69.77 | 100.00 | 99.17 |
| | | | pltn | 73.36 | 14.37 | 39.76 | .4510 | 78.35 | 100.00 | 99.26 |
| HW-TSC | U | prmry | ALL | 65.41 | 25.29 | 50.38 | .5579 | 72.10 | 59.42 | 100.00 |
| | | | ted | 44.50 | 44.83 | 68.02 | .7326 | 82.83 | 31.93 | 100.00 |
| | | | itv | 68.20 | 22.60 | 46.72 | .5319 | 68.95 | 67.58 | 100.00 |
| | | | pltn | 74.95 | 17.21 | 44.07 | .5152 | 73.94 | 54.50 | 100.00 |
| HW-TSC | U | cntrstv1 | ALL | 65.97 | 25.21 | 50.49 | .5612 | 33.25 | 66.05 | 100.00 |
| | | | ted | 44.45 | 44.63 | 68.08 | .7353 | 48.76 | 38.26 | 100.00 |
| | | | itv | 69.27 | 22.56 | 46.84 | .5338 | 25.02 | 72.91 | 100.00 |
| | | | pltn | 74.95 | 17.19 | 44.22 | .5213 | 44.16 | 64.58 | 100.00 |
| FBK-AI4C$_{CSC}$ | U | prmry | ALL | 66.02 | 23.87 | 46.53 | .4811 | 67.56 | 94.25 | 100.00 |
| | | | ted | 40.81 | 43.11 | 68.20 | .7408 | 81.79 | 92.20 | 100.00 |
| | | | itv | 71.62 | 19.18 | 39.70 | .4019 | 62.11 | 94.22 | 100.00 |
| | | | pltn | 73.16 | 16.19 | 42.78 | .4921 | 69.30 | 95.60 | 100.00 |
| APPTEK | U | cntrstv2 | ALL | 68.69 | 19.83 | 45.46 | .4817 | 71.43 | 100.00 | 99.00 |
| | | | ted | 48.14 | 35.78 | 62.51 | .6681 | 82.76 | 100.00 | 97.74 |
| | | | itv | 71.58 | 17.85 | 42.21 | .4572 | 66.60 | 100.00 | 99.25 |
| | | | pltn | 77.76 | 12.62 | 38.75 | .4301 | 75.54 | 100.00 | 99.14 |
| FBK-AI4C$_{DIR}$ | C | prmry | ALL | 70.09 | 19.16 | 41.58 | .3972 | 73.08 | 91.64 | 99.97 |
| | | | ted | 40.45 | 42.09 | 67.76 | .7224 | 82.59 | 89.77 | 99.93 |
| | | | itv | 78.20 | 12.09 | 31.50 | .2827 | 70.11 | 92.89 | 100.00 |
| | | | pltn | 75.52 | 13.20 | 40.33 | .4389 | 72.01 | 90.84 | 99.96 |
| Submissions 2023 (here ALL={ted,itv,pltn}, while last year *eptv* was considered as well): | | | | | | | | | | |
| MATESUB | U | prmry | ALL | 67.29 | 22.54 | 46.40 | .4993 | 85.51 | 99.53 | 100.00 |
| APPTEK | C | prmry | ALL | 72.33 | 17.72 | 38.49 | .3467 | 95.30 | 100.00 | 100.00 |
| FBK | C | prmry | ALL | 73.93 | 16.70 | 37.68 | .3217 | 76.57 | 91.84 | 99.99 |

Table 38: Subtitling Task: automatic evaluation scores on tst2023 en→es. *C* and *U* stand for *constrained* and *unconstrained* training condition, respectively; *prmry* and *cntrstv* for *primary* and *contrastive* systems. Ranking based on SubER scores on ALL domains.

## B.4 Speech-to-Speech Translation

| System | D | Test | | |
|---|---|---|---|---|
| Ref | | BLEU | chrF | COMET |
| *Cascade Systems* | | | | |
| HW-TSC | U | 33.6 | 29.4 | 74.79 |
| | C$^+$ | 31.8 | 28.1 | 74.41 |
| | C | 31.4 | 28.5 | 73.65 |

Table 39: Official results of the **automatic evaluation** for the English to Chinese Speech-to-Speech Translation Task. The "D" column indicates the data condition in which each submitted run was trained, namely: Constrained (C), Constrained$^{+LLM}$ (C$^+$), Unconstrained (U).

## B.5 Low-Resource SLT

**North Levantine Arabic→English (Unconstrained Condition)**

| Team | System | BLEU↓ | chrF2 | COMET |
|---|---|---|---|---|
| ALADAN | primary | 28.71 | 52.25 | 0.7763 |
| ALADAN | contrastive1 | 28.50 | 52.12 | 0.7706 |
| ALADAN | contrastive2 | 22.12 | 46.38 | 0.7296 |
| KIT | primary | 20.86 | 44.54 | 0.7013 |
| KIT | contrastive1 | 19.73 | 45.43 | 0.7098 |
| JHU | primary | 15.95 | 38.89 | 0.6951 |
| JHU | contrastive1 | 14.74 | 37.27 | 0.6775 |
| HW-TSC | primary | 13.64 | 33.31 | 0.5877 |
| KIT | contrastive2 | 11.87 | 34.76 | 0.6064 |
| UM | contrastive1 | 5.09 | 24.50 | 0.5378 |
| UM | primary | 4.74 | 24.10 | 0.5369 |
| UM | contrastive2 | 3.53 | 21.56 | 0.5196 |

Table 40: Automatic evaluation results for the North Levantine Arabic to English task, unconstrained Condition. A lowercase, no punctuation variant of chrF2 is reported. The `Unbabel/wmt22-comet-da` model was used for COMET computation, with the source side (Arabic transcript) unmodified and the target side lowercased and with removed punctuation.

**Bemba→English (Unconstrained Condition)**

| Team | System | BLEU |
|---|---|---|
| JHU | primary | 32.6 |
| KIT | primary | 28.8 |
| KIT | contrastive2 | 28.1 |
| JHU | contrastive1 | 27.0 |
| KIT | contrastive1 | 27.0 |
| JHU | contrastive2 | 26.7 |

| Team | System | WER |
|---|---|---|
| KIT ASR | primary | 33.2 |
| JHU ASR | primary | 35.7 |

Table 41: Automatic evaluation results for the Bemba to English task, unconstrained Condition.

**Bhojpuri→Hindi (Unconstrained Condition)**

| Team | System | BLEU | chrF2 |
|---|---|---|---|
| JHU | primary | 24.4 | 49.5 |
| JHU | contrastive1 | 23.9 | 48.7 |
| JHU | contrastive2 | 12.2 | 39.1 |
| BITSP | primary | 12.9 | 41.1 |
| DFKI_MLT | primary | 0.1 | 6.1 |

Table 42: Automatic evaluation results for the Bhojpuri to Hindi task, unconstrained Condition.

**Irish→English (Unconstrained Condition)**

| Team | System | BLEU | chrF2 |
|---|---|---|---|
| JHU | contrastive1 | 16.0 | 39.0 |
| JHU | primary | 15.3 | 38.3 |
| Ymoslem | primary | 7.6 | 27.6 |
| Ymoslem | contrastive1 | 7.4 | 26.5 |
| Ymoslem | contrastive2 | 5.1 | 24.7 |
| SETU-DCU | primary | 0.6 | 15.4 |

Table 43: Automatic evaluation results for the Irish to English task, unconstrained Condition.

**Maltese→English (Unconstrained Condition)**

| Team | System | BLEU | chrF2 |
|---|---|---|---|
| KIT | primary | 58.9 | 76.5 |
| SETU-DCU | primary | 56.7 | 81.9 |
| KIT | contrastive2 | 56.2 | 75.0 |
| KIT | contrastive1 | 55.2 | 74.4 |
| SETU-DCU | contrastive1 | 52.6 | 72.1 |
| UoM | primary | 52.4 | 72.3 |
| UoM | contrastive1 | 52.4 | 72.3 |
| UoM | contrastive2 | 52.3 | 72.1 |
| SETU-DCU | contrastive2 | 44.7 | 65.5 |
| JHU | primary | 41.4 | 68.6 |
| JHU | contrastive1 | 36.5 | 64.2 |
| UoM-DFKI | primary (e2e) | 35.1 | 59.0 |
| JHU | contrastive2 | 24.8 | 55.8 |
| UoM-DFKI | contrastive1 (e2e) | 18.5 | 42.0 |

Table 44: Automatic evaluation results for the Maltese to English task, Unconstrained Condition. e2e denotes end-to-end system.

**Maltese→English (Constrained Condition)**

| Team | System | BLEU | chrF2 |
|---|---|---|---|
| UoM | primary | 0.5 | 15.6 |

Table 45: Automatic evaluation results for the Maltese to English task, Constrained Condition.

**Marathi→Hindi (Unconstrained Condition)**

| Team | System | BLEU | chrF2 |
|---|---|---|---|
| IITM | primary | 47.2 | 70.1 |
| JHU | primary | 37.7 | 62.7 |
| JHU | contrastive1 | 37.3 | 62.4 |
| JHU | contrastive2 | 28.5 | 55.0 |
| BITSP | contrastive1 | 25.0 | 50.1 |
| BITSP | primary | 21.3 | 48.1 |
| BITSP | contrastive2 | 19.0 | 44.8 |

| Team | System | WER | CER |
|---|---|---|---|
| IITm ASR | primary | 22.8 | 7.3 |
| JHU ASR | primary | 26.7 | 8.9 |
| BITSP ASR | contrastive1 | 62.9 | 17.5 |
| BITSP ASR | primary | 69.3 | 21.2 |
| BITSP ASR | contrastive2 | 69.3 | 21.2 |

Table 46: Automatic evaluation results for the Marathi to Hindi task, Unconstrained Condition.

**Quechua→Spanish (Constrained Condition)**

| Team | System | BLEU | chrF2 |
|---|---|---|---|
| QUESPA | contrastive2 | 1.3 | 30.9 |
| QUESPA | contrastive1 | 1.4 | 30.3 |
| QUESPA | primary | 2.0 | 30.0 |

Table 47: Automatic evaluation results for the Quechua to Spanish task, Constrained Condition. ChrF2 scores were only taken into account for those systems that scored less than 5 points BLEU.

**Quechua→Spanish (Unconstrained Condition)**

| Team | System | BLEU | chrF2 |
|---|---|---|---|
| QUESPA | contrastive1 | 19.7 | 43.1 |
| QUESPA | primary | 16.0 | 52.2 |
| JHU | primary | 12.5 | 49.7 |
| QUESPA | contrastive2 | 11.1 | 44.6 |
| JHU | contrastive1 | 6.4 | 39.5 |
| JHU | contrastive2 | 0.9 | 13.0 |

Table 48: Automatic evaluation results for the Quechua to Spanish task, Unconstrained Condition.

**Tamasheq→French (Unconstrained Condition)**

| Team | System | BLEU |
|---|---|---|
| Organizer Baseline | primary | 8.83 |
| JHU | primary | 6.07 |
| JHU | contrastive | 0.50 |

Table 49: Automatic evaluation results for the Tamasheq to French task, Unconstrained Condition.